\documentclass[10pt,twocolumn,letterpaper]{article}

\usepackage{iccv}
\usepackage{times}
\usepackage{epsfig}
\usepackage{graphicx}
\usepackage{amsmath}
\usepackage{amssymb}
\usepackage{multirow}
\usepackage{booktabs}
\usepackage{subfigure}

\usepackage[pagebackref=true,breaklinks=true,letterpaper=true,colorlinks,bookmarks=false]{hyperref}
\usepackage{bbding}

\newcommand\blfootnote[1]{%
  \begingroup
  \renewcommand\thefootnote{}\footnote{#1}%
  \addtocounter{footnote}{-1}%
  \endgroup
}

\iccvfinalcopy 

\def\iccvPaperID{***} 

\ificcvfinal\fi

\begin{document}

\title{MultiSports: A Multi-Person Video Dataset of Spatio-Temporally Localized Sports Actions}

\author{Yixuan Li \qquad Lei Chen \qquad Runyu He \qquad Zhenzhi Wang \qquad Gangshan Wu\qquad Limin Wang\textsuperscript{\Envelope} \\
State Key Laboratory for Novel Software Technology, Nanjing University, China\\}
\maketitle
\ificcvfinal\thispagestyle{empty}\fi

\begin{abstract}
   Spatio-temporal action detection is an important and challenging problem in video understanding. The existing action detection benchmarks are limited in aspects of small numbers of instances in a trimmed video or low-level atomic actions. This paper aims to present a new multi-person dataset of spatio-temporal localized sports actions, coined as {\em MultiSports}. We first analyze the important ingredients of constructing a realistic and challenging dataset for spatio-temporal action detection by proposing three criteria: (1) multi-person scenes and motion dependent identification, (2) with well-defined boundaries, (3) relatively fine-grained classes of high complexity. Based on these guidelines, we build the dataset of MultiSports v1.0 by selecting 4 sports classes, collecting 3200 video clips, and annotating 37701 action instances with 902k bounding boxes. Our datasets are characterized with important properties of high diversity, dense annotation, and high quality. Our MultiSports, with its realistic setting and detailed annotations, exposes the intrinsic challenges of spatio-temporal action detection. To benchmark this, we adapt several baseline methods to our dataset and give an in-depth analysis on the action detection results in our dataset. We hope our MultiSports can serve as a standard benchmark for spatio-temporal action detection in the future. Our dataset website is at \href{https://deeperaction.github.io/multisports/}{https://deeperaction.github.io/multisports/}.
\end{abstract}
\blfootnote{ \Envelope: Corresponding author (lmwang@nju.edu.cn).}

\section{Introduction}
\begin{figure*}
\begin{center}
\includegraphics[width=17.5cm]{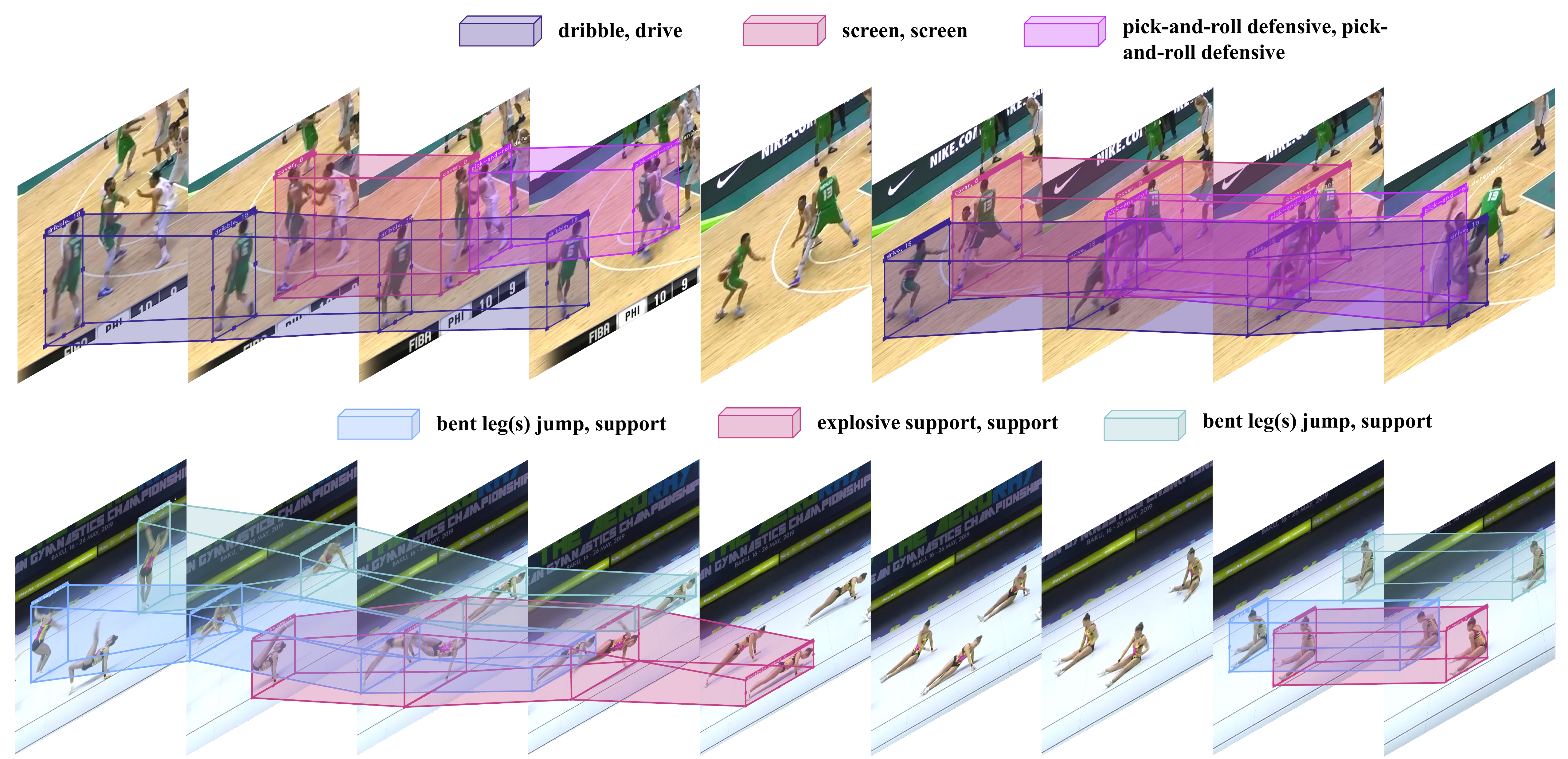}
\end{center}
\vspace{-1em}
   \caption{The 25fps tubelets of bounding boxes and fine-grained action category annotations in MultiSports dataset. Multiple concurrent action situations frequently appear in MultiSports with many starting and ending points in the long untrimmed video clips. The frames are cropped and sampled by stride 5 or 7 for visualization propose. Tubes with the same color represent the same person.}
\label{fig:first}
\vspace{-1em}
\end{figure*}

Spatio-temporal human action detection in untrimmed videos is of great importance for many applications, such as surveillance and sports analysis. Recently, recognizing actions from short trimmed videos has achieved considerable progress~\cite{DBLP:conf/eccv/WangXW0LTG16,DBLP:conf/cvpr/CarreiraZ17,DBLP:conf/iccv/TranBFTP15,DBLP:conf/nips/SimonyanZ14,DBLP:conf/cvpr/WangL0G18,DBLP:conf/cvpr/Wang0T15}, but these classification models can not be directly applied for video analysis in a multi-person scene. Meanwhile, although temporal action detection methods~\cite{DBLP:conf/iccv/ZhaoXWWTL17,DBLP:conf/eccv/LinZSWY18,DBLP:conf/iccv/LinLLDW19,DBLP:conf/iccv/XuDS17,DBLP:conf/iccv/ZengHGTRZH19} for untrimmed videos can distinguish intervals of human actions from background, they are still unable to spatially detect multiple concurrent human actions, which is important in real-world applications of video analysis.

Current spatio-temporal action detection benchmarks can be mainly classified into two categories: 1) Densely annotated high-level actions such as J-HMDB~\cite{DBLP:conf/iccv/JhuangGZSB13} and UCF101-24~\cite{DBLP:journals/corr/abs-1212-0402}. Their clips only have a single person doing some semantically simple and temporally repeated actions. Typically, the scene context can provide enough cues for recognizing these coarse-grained action categories. Thus, these benchmarks might be impractical for real-world applications such as surveillance, where it is required to deal with more fine-grained actions in a multi-person scene; 2) Sparsely annotated atomic actions such as AVA~\cite{DBLP:conf/cvpr/GuSRVPLVTRSSM18}. They fail to provide clear temporal action boundaries, and simply focus on frame-level spatial localization of atomic actions. This setting removes the requirements of temporal localization for action detection algorithms. Meanwhile, their atomic actions rarely require the complex reasoning over the actors and their surrounding environment.

Based on the analysis above, we argue that a new benchmark is necessary to advance the research of spatio-temporal action detection. The benchmark should satisfy several important requirements to cover the realistic challenges of this task. 1) There should be multiple persons performing different actions concurrently in the same scene, where the background information is not sufficient for action recognition and motion itself of the actor plays a significant role. 2) To address the inherently confusing human action boundaries in time, actions should be both semantically and temporally well-defined with a consensus among humans. 3) Considering the complexity of real-world applications, actions should be fine-grained which requires accurate human pose and motion information, long-term temporal structure, possible interactions between humans, objects and scenes, and reasoning over their relations.

Following the above guidelines, we develop the {\it MultiSports} dataset, short for {\it Multi-person Sports Actions}. The dataset is large-scale, high-quality, multi-person, and contains fine-grained action categories with precise and dense annotations in both spatial and temporal domains. The action vocabulary consists of 66 action classes collected from 4 sports (basketball, volleyball, football and aerobic gymnastics). An example clip has been visualized in Figure~\ref{fig:first}. We choose these four sports for the following reasons. 1) There are plenty of multiple concurrent action instances in sports competitions. Also, the background is far less characteristic and cannot provide sufficient information for fine-grained action recognition. 2) Sports actions have well-defined categories and boundaries. These boundaries are defined by either professional athletes or official documentations~\cite{de2017aerobic}. 3) Due to the complex competition rules, recognizing sports action generally requires to model the long-term structure and the human-object-scene interactions. For example, in football, although the athlete may take only 0.5s to kick the ball, we may need up to 5s context to recognize whether it is pass, long ball, through ball, or cross. 


In practice, we conduct exhaustive annotations of 25 fps frame-wise bounding boxes and fine-grained action categories in a two-stage procedure: 1) a team of professional athletes of corresponding sport to annotate the temporal and category labels, and 2) a team of crowd-sourced annotators to finish the bounding boxes with the help of tracking method FCOT~\cite{DBLP:journals/corr/abs-2004-07109}. This two-stage annotation procedure as well as careful quality control together can guarantee consistent and clean annotations. 
To ensure the visual quality, all videos in our dataset are high-resolution records of professional competitions from a diversity of countries and different performance levels. 

Given the well-defined and dense-annotated action instances in {\it MultiSports v1.0}, we benchmark spatio-temporal action detection on this challenging dataset. We perform empirical studies with several recent state-of-the-art action detector methods. Compared with previous action detection benchmarks such as J-HMDB~\cite{DBLP:conf/iccv/JhuangGZSB13} and UCF101-24~\cite{DBLP:journals/corr/abs-1212-0402}, our MultiSports is quite challenging with a much lower frame mAP and video mAP. We also introduce a detailed error analysis on detection results and try to provide more insights on spatio-temporal action detection.  According to our analysis on MultiSports benchmark, we figure out several challenges of spatio-temporal action detection that needs to be addressed, such as capturing subtle differences between fine-grained action categories, performing accurate temporal localization, dealing with action occlusion and modeling long-range context. We hope MultiSports could serve as a standard benchmark to advance the area of spatio-temporal action detection in the future. MultiSports sptatio-temporal action detection is currently a track of DeeperAction challenge at ICCV 2021 \href{https://deeperaction.github.io/}{https://deeperaction.github.io/}.

In summary, our main contribution is twofold. 1) We develop a new benchmark MultiSports of spatio-temporal action detection for well-defined and realistically difficult human actions in a multi-person scene, providing high-quality and 25fps frame-wise annotations from four sports. 2) We conduct extensive studies and systematic error analysis on MultiSports, which reveals the key challenges of spatio-temporal action detection and  hopefully can facilitate future research in this area.

\section{Related Work}
\noindent{\bf Action recognition datasets.} Early datasets of action recognition mainly focus on action classification. Those datasets, including KTH~\cite{DBLP:conf/icpr/SchuldtLC04}, Weizmann~\cite{DBLP:conf/iccv/BlankGSIB05}, UCF-101~\cite{DBLP:journals/corr/abs-1212-0402} and HMDB~\cite{DBLP:conf/iccv/KuehneJGPS11}, contains manually trimmed short clips to capture semantics of a single action. Their human action cues, however, are overwhelmed by signals of background scenes. Multi-MiT~\cite{DBLP:journals/corr/abs-1911-00232} is a multi-label action recognition dataset, which may have several concurrent actions but do not provide temporal duration and spatial annotations. Recently, large-scale video classification datasets such as Sports-1M~\cite{DBLP:conf/cvpr/KarpathyTSLSF14}, YouTube-8M~\cite{DBLP:journals/corr/Abu-El-HaijaKLN16} and Kinetics~\cite{DBLP:conf/cvpr/CarreiraZ17} have been created for feature representation learning and serve as pre-training in downstream tasks, but appearance cues still play a important role here.  Something-something~\cite{DBLP:conf/iccv/GoyalKMMWKHFYMH17} and FineGym~\cite{DBLP:conf/cvpr/ShaoZDL20a}, with plenty of fine-grained action categories, effectively reduce the influences of background scenes and reveal some key challenges of modeling a single action. They share the similar property of capturing motion cues with {\it MultiSports}, but only have one concurrent action therefore we address a different need with them.

Temporal action detection datasets such as ActivityNet~\cite{DBLP:conf/cvpr/HeilbronEGN15}, HACS~\cite{DBLP:conf/iccv/Zhao0TY19}, THUMOS14~\cite{DBLP:journals/cviu/IdreesZJGLSS17}, MultiTHUMOS~\cite{DBLP:journals/ijcv/YeungRJAMF18} and Charades~\cite{DBLP:conf/eccv/SigurdssonVWFLG16} provide temporal action detection annotations for each action of interest in untrimmed videos. But unlike MultiSports, they do not provide spatial annotations and could not identify multiple concurrent actions for multiple people. 

Previous spatio-temporal action detection datasets, such as UCF Sports~\cite{DBLP:conf/cvpr/RodriguezAS08}, UCF101-24~\cite{DBLP:journals/corr/abs-1212-0402} and J-HMDB~\cite{DBLP:conf/iccv/JhuangGZSB13}, typically evaluate spatio-temporal action detection for short videos with only a single person and coarse-grained action categories. Our MultiSports significantly differs from them in several aspects: multiple concurrent actions by multiple people; less characteristic background scenes; the larger number of action and fine-grained categories; more fast movement and large deformation; and significantly more instances per clip. Recently, a new type of extensions such as DALY~\cite{DBLP:journals/corr/WeinzaepfelMS16}, AVA~\cite{DBLP:conf/cvpr/GuSRVPLVTRSSM18} and AVA-Kinetics~\cite{DBLP:journals/corr/abs-2005-00214} adopt sparse annotations of daily life actions, either in composite or atomic forms, to reduce human labors of annotating and increase the scale of datasets. It may be a good way for evaluating daily life actions without fast movement and large deformation, but unsuitable for areas like sports analysis, since it often requires continuous annotations of all human actions of interest. MEVA~\cite{DBLP:conf/wacv/CoronaOCH21} is a security dataset, which provides spatial-temporal annotations and some other modality annotations. But our sports actions are more complex and fast-changing than MEVA. Different from previous datasets, our MultiSports proposes a more difficult benchmark with multi-person, well-defined boundaries, fine-grained setting and frame-by-frame annotations, which focuses on the sports domain.

\noindent{\bf Spatio-temporal action detection.} Most recent approaches for UCF101-24 and JHMDB can be classified into two categories: frame-level detectors and clip-level detectors. Many efforts have been made to extend an image object detector to the task of spatio-temporal action detection at the frame level~\cite{DBLP:conf/cvpr/GkioxariM15,DBLP:conf/cvpr/Wang0TG16,DBLP:conf/eccv/PengS16,DBLP:conf/bmvc/SahaSSTC16,DBLP:conf/iccv/SinghSSTC17,DBLP:conf/iccv/WeinzaepfelHS15}, where the resulting per-frame detections are then linked to generate final tubes. Although flows could be used to capture motion cues, frame-level detector fails to fully utilize temporal information. To model temporal structures for action detection, some clip-level approaches or action tubelet detectors~\cite{DBLP:conf/iccv/HouCS17,DBLP:conf/eccv/LiQDYM18,DBLP:conf/iccv/KalogeitonWFS17a,DBLP:conf/cvpr/YangY0XDK19,DBLP:conf/eccv/LiW0W20,DBLP:conf/cvpr/ZhaoS19,DBLP:conf/cvpr/SongZY019} have been proposed. ACT~\cite{DBLP:conf/iccv/KalogeitonWFS17a} took several frames as input and detected tubelets regressed from anchor cuboids. STEP~\cite{DBLP:conf/cvpr/YangY0XDK19} progressively refined the proposals by a few steps to solve the large displacement problem and utilized longer temporal information. MOC-detector~\cite{DBLP:conf/eccv/LiW0W20} proposed an anchor-free tubelet detector by treating action instances as trajectories of moving points. For AVA, many methods~\cite{DBLP:conf/iccv/Feichtenhofer0M19,DBLP:conf/cvpr/GirdharCDZ19, DBLP:conf/eccv/TangXMPL20, DBLP:conf/cvpr/WuF0HKG19, DBLP:conf/eccv/WuKWZW20} have been proposed to better make use of spatio-temporal information for atomic action classification.

\section{The MultiSports Dataset}
\begin{figure*}[t]
\begin{center}
\includegraphics[width=17.5cm]{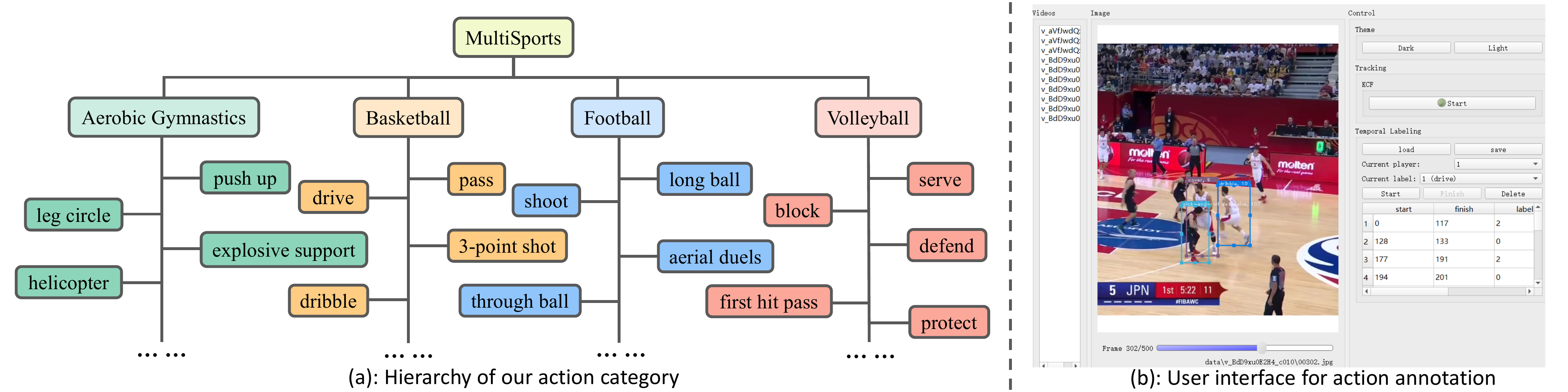}
\end{center}
\vspace{-1em}
   \caption{The action vocabulary hierarchy and annotator interface of the MultiSports dataset. (a) Our MultiSports has a two-level hierarchy of action vocabularies, where the actions of each sport are fine-grained. (b) Details of annotations can be found in Sec~\ref{sec:anno}.}
\label{fig:hierarchy}
\vspace{-1em}
\end{figure*}

Our {\it MultiSports} dataset aims to introduce a new challenging benchmark with high-quality annotations to the area of spatio-temporal action detection, which differs from previous ones in multi-person scene, well-defined temporal boundaries, and fine-grained action categories. Sec.~\ref{sec:anno} introduces our annotation procedure. Statistics and characteristics of MultiSports are elaborated in Sec.~\ref{sec:stat} and Sec.~\ref{sec:chara}.

\subsection{Dataset Construction}
\label{sec:anno}
\noindent{\bf Action vocabulary generation.} We select sports of basketball, volleyball, football and aerobic gymnastics, because of their multi-person setting, less ambiguous actions and well-defined temporal boundary. For aerobic gymnastics, we use the official documentations~\cite{de2017aerobic}. In practice, we only select {\it difficulty elements} and discard {\it movement patterns}. For the remaining ball sports, we use an iterative way to generate our action vocabulary in each sport: we initialize an action list by the suggestions of athletes and write a handbook to clarify the definition of action boundaries. Then we let several annotators try to annotate the data, where inaccurate definitions of action boundaries, ambiguities between action categories and missed action categories will be collected from their feedback. We iteratively adjust our action list and handbook according to the feedback several times before we start massive annotating, which results in the final action hierarchy shown in Figure.~\ref{fig:hierarchy}(a). Note that the annotators of action categories and temporal boundaries are professional athletes of the corresponding sports, so their feedback is important for building a well-defined action vocabulary in practice. To keep action boundaries accurate and make our dataset suitable for spatio-temporal action detection, we do not count common and atomic actions such as run or stand in our action vocabulary. We also exclude foul in ball sports. Because in the 2D video records, we recognize fouls most from the referee's reaction instead of the actor's motion. What is worse, it is hard to identify who fouls due to occlusion.

\noindent{\bf Data preparation.} After choosing the four sports, we search for their competition videos by querying the name of sports like volleyball and the name of competition levels like Olympics and World Cup on YouTube, and then download videos from top search results. For each video, we only select high-resolution, {\it e.g.} 720P or 1080P, competition records and then manually cut them into clips of minutes, with less shot changes in each clip and to be more suitable for action detection. These official records share consistent and rich content, and can guarantee a high-quality dataset.

\noindent{\bf Action annotation.} Since our annotations are difficult in labeling fine-grained categories and exhaustive in determining 25fps frame-wise bounding boxes, we naturally decompose our annotation procedure into two stages: 1) A team of professional athletes generate records of the action label, the starting and ending frame, and the person box in the starting frame, which can ensure the efficiency, accuracy and consistency of our annotation results; 2) With the help of FCOT~\cite{DBLP:journals/corr/abs-2004-07109} tracking algorithms, a team of crowd-sourced annotators adjust bounding boxes of tracking results at each frame for each record. The ambiguity of spatial human boundaries is much less than that of fine-grained action categories and temporal action boundaries. They use the interface shown in Figure~\ref{fig:hierarchy}(b). 

To ensure the consistency of action temporal boundaries, which tends to be ambiguous and remains as a big challenge for most temporal action detection datasets, we write a handbook to clarify the definition of action boundaries as mentioned above. For example, our handbook unifies the annotations of {\it football pass} as starting from the ball-controlling-leg leaving the ground and ending with this leg touching the ground again. The annotation handbook is provided in Appendix E.

\noindent{\bf Person bounding-box tracking.} As mentioned above, we first tack each record generated by professional athletes and then employ crowd-sourced annotators to refine the bounding boxes at each frame. Specifically, we use FCOT~\cite{DBLP:journals/corr/abs-2004-07109} to track the bounding boxes frame-by-frame. We find this tracking-to-refinement labeling process can not only speed up the annotation process, but also increase the annotation quality by enforcing workers to focus on  determining precise boundary of each box.

We also evaluate the output of FCOT~\cite{DBLP:journals/corr/abs-2004-07109} and results are shown in Table~\ref{tab:tracking_res}. We adopt success and precision metrics proposed in OTB100~\cite{DBLP:journals/pami/WuLY15}. Aerobic turned out the hardest in both success and precision aspects.

\begin{table}[htbp]
    \footnotesize
    \centering
    \begin{tabular}{l|cccc}
    \toprule[1pt]
         & Aerobic gym. & Volleyball & Football & Basketball \\
        \hline
        Success     & 0.66 & 0.72 & 0.77 & 0.66\\
        Precision   & 0.67 & 0.93 & 0.92 & 0.72\\
    \bottomrule[1pt]
    \end{tabular}
    \vspace{0.5em}
    \caption{Tracking results on different sports}
    \vspace{-1em}
    \label{tab:tracking_res}
\end{table}

\noindent{\bf Quality control.} For the first stage of annotation, every clip has at least one annotator with domain knowledge double-checking the annotations. We correct wrong or inaccurate ones and also add missing annotations for a higher recall, e.g., adding missed defence action in football and modifying inconsistent action boundaries. For the second stage, we double-check each instance by playing it in 5fps and manually correct the inaccurate bounding boxes.

\subsection{Dataset Statistics}
\label{sec:stat}

\begin{table}[t]
	\footnotesize
	\centering
	 \resizebox{0.48\textwidth}{!}{
	\begin{tabular}{c||c|c|c|c|c}
	\toprule[1pt]
		& anno type & \# act.   & \# inst.  & avg act./vid. dur. & \# bbox\\ 
		\midrule[1pt]
		 J-HMDB~\cite{DBLP:conf/iccv/JhuangGZSB13} & Tube & 21  &  928  &  1.2s / 1.2s  &	32k\\ 
		UCF101-24~\cite{DBLP:journals/corr/abs-1212-0402}& Tube &  24  &   4458  &   5.1s / 6.9s  & 574k\\ 
		AVA V2.1~\cite{DBLP:conf/cvpr/GuSRVPLVTRSSM18}$^\ast$& Frame &  80      &  \textasciitilde 56000$^\dag$  &   Sparse$^\ddag$ /15m  & 426k\\
		AVA-Kinetics~\cite{DBLP:journals/corr/abs-2005-00214}$^\ast$& Frame &  80      &  \textasciitilde 186000$^\dag$  &   -  & 590k\\
		HACS~\cite{DBLP:conf/iccv/Zhao0TY19}& Segment & 200 & 140k & 33.2s / 148.7s &-\\
		FineGym V1.0~\cite{DBLP:conf/cvpr/ShaoZDL20a}& Segment & 530  & 32697  &   1.7s / 10m  & -\\ 
		\midrule[0.9pt]
		Aerobic gym. & Tube    &   21           &   8703  &  1.5s / 30.7s  &   325k \\ 
		Volleyball   & Tube    &  12          &  7645  &  0.7s / 10.5s    & 139k   \\  
		Football & Tube 	& 	15	  &  12254	   &  0.7s / 22.6s  &  225k \\ 
		Basketball & Tube & 	18	  &  9099	   &  0.9s / 19.7s  &  213k \\ 
		\midrule[0.6pt]
		Ours in total & Tube & 	66	  &  37701	   &  1.0s / 20.9s  &  902k \\ 
		\bottomrule[1pt]
	\end{tabular}
	 }
	\vspace{0.8mm}
	\caption{Comparison of statistics between existing action detection datasets and our MultiSports v1.0. ($^\ast$ only train and val sets' ground-truths are available; $Tube$ with class, temporal boundary and spatial localization; $Frame$ with class and spatial localization; $Segment$ with class and temporal boundary; $^\dag$ number of person tracklets, each of which has one or more action labels; $^\ddag$ 1fps action annotations)}
	\vspace{-2em}
	\label{tab:stat}
\end{table}

\begin{figure*}[t]
\begin{center}
\includegraphics[width=17.5cm]{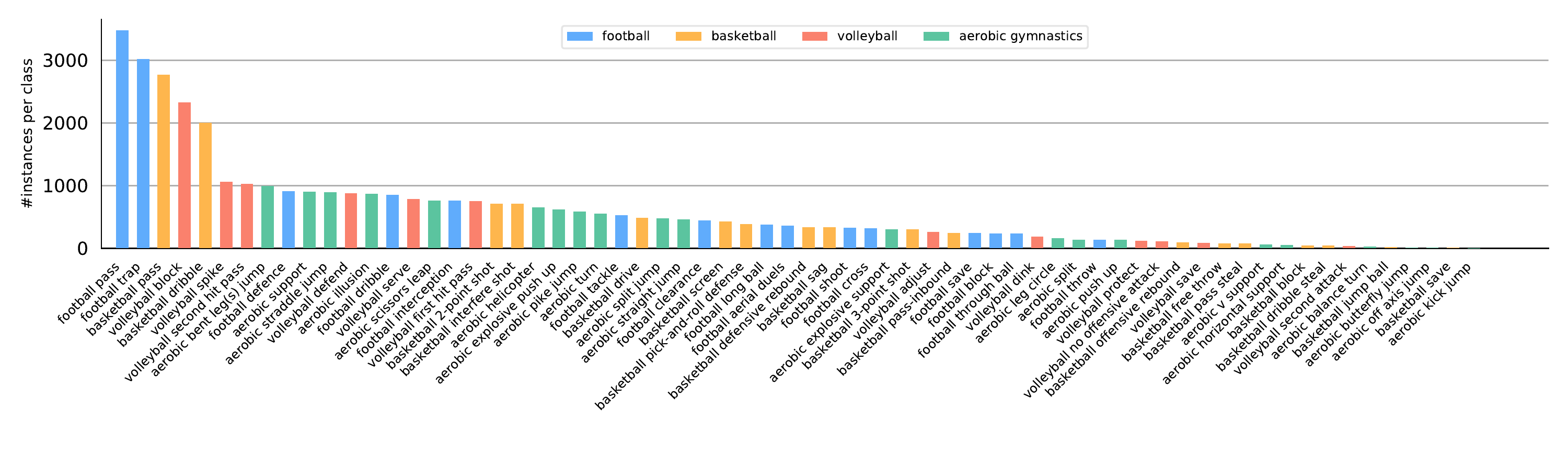}
\end{center}
\vspace{-2em}
   \caption{Statistics of each action class's data size in MultiSports, which is sorted by descending order with 4 colors indicating 4 different sports. For actions in the different sports sharing the same name, we add the name of sports before them.}
\label{fig:num_instance}
\vspace{-1em}
\end{figure*}

Our {\it MultiSports} v1.0 contains 66 fine-grained action categories from four sports, and has videos selected from 247 competitions. The videos are manually cut into 800 clips per sport to keep data balance between sports. We discard intervals with only background scenes, such as award, and select the highlights of competitions as clips for action detection. Table~\ref{tab:stat} compares the annotation types and statistics of MultiSports v1.0 with the existing datasets.
AVA~\cite{DBLP:conf/cvpr/GuSRVPLVTRSSM18} only has sparse and 1fps annotations of bounding boxes, which fails to provide clear temporal action boundaries and focuses on atomic action recognition. AVA-Kinetics~\cite{DBLP:journals/corr/abs-2005-00214} uses part of 10s clips of the Kinetics~\cite{DBLP:conf/cvpr/CarreiraZ17} and annotates one key frame per clip without any temporal boundary annotations either. Our annotation type is different from theirs.
MultiSports distinguishes with existing datasets such as J-HMDB~\cite{DBLP:conf/iccv/JhuangGZSB13} and UCF101-24~\cite{DBLP:journals/corr/abs-1212-0402} in longer untrimmed video clips (20.9s vs. 1.2s or 6.9s), more fine-grained action categories (66 vs. 21 or 24), much more instances (37701 vs. 928 or 4458), and more instances per video clip (11.8 vs. 1 or 1.4), which raises new challenges of modeling fast movement and fine-grained actions of multiple people in a longer video. Our MultiSports also has the largest number of bounding boxes among all existing datasets. We find that fine-grained category and well-defined boundary usually greatly shorten the action duration, which agrees with FineGym~\cite{DBLP:conf/cvpr/ShaoZDL20a}. Also, we only keep the common part of actions in ball sports for well-defined boundaries. For instance, {\it basketball pass} starts from the player pushing the ball outwards with his arms, but does not include holding the ball and doing fake actions. Therefore our average action duration is smaller than UCF101-24 and HACS~\cite{DBLP:conf/iccv/Zhao0TY19}, which contains coarse-grained and temporally repeated actions such as {\it volleyball} in HACS and {\it riding horses} in UCF101-24.

\begin{figure}[t]
\begin{center}
\includegraphics[width=8cm]{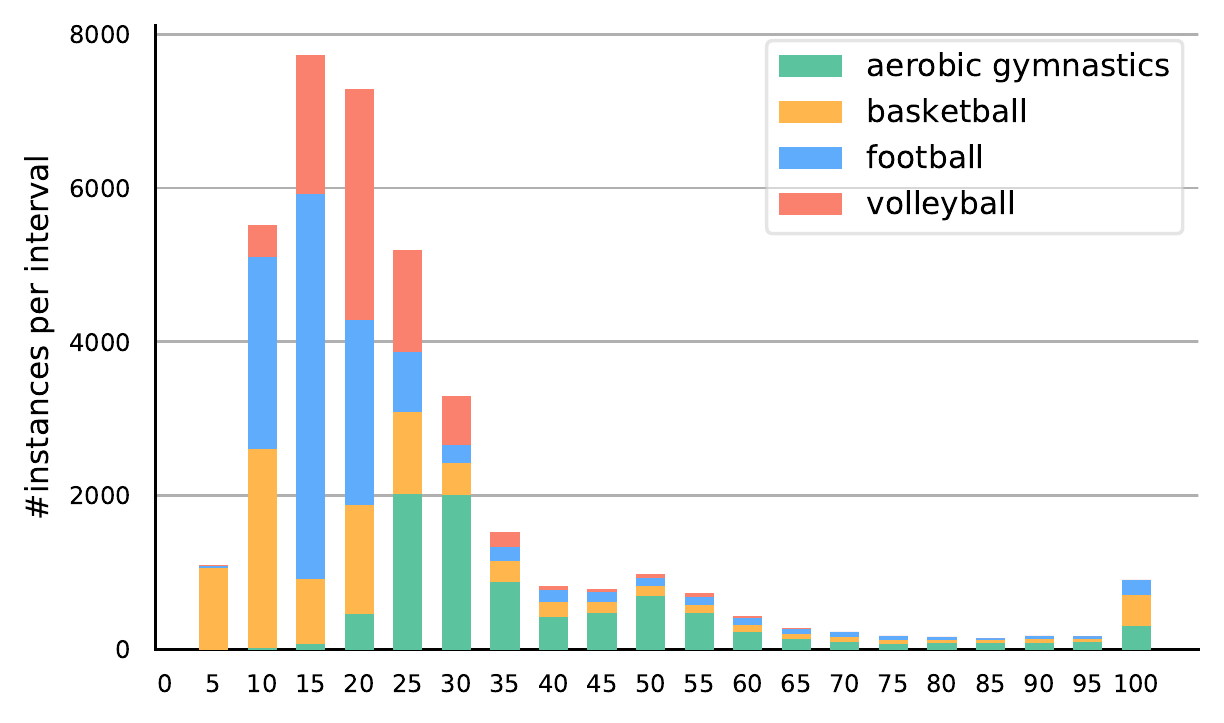}
\end{center}
\vspace{-1.5em}
   \caption{Statistics of action instance duration in MultiSports, where the x-axis is the number of frames and we count all instances longer than 95 frames in the last bar.}
\label{fig:interval}
\vspace{-1.5em}
\end{figure}

As shown in Figure~\ref{fig:num_instance}, the instance number of each action category ranges from 3 to 3,477,  showing the natural long-tailed distribution~\cite{DBLP:journals/corr/abs-1709-01450}. The long-tailed action categories also raise new challenges for action detection models. Figure~\ref{fig:interval} shows the distribution of action instance duration. The large variations of action instance duration add more difficulty for action detection models to accurately localize temporal boundary. Moreover, action instances in MultiSports are often related with longer temporal context and interactions with context. These inherent challenges of MultiSports require a more powerful and flexible temporal modeling scheme for action detection.

Our training/validation/test sets are split at the clip level, where the clip numbers in each sport are manually controlled as 3:1:2 for training/validation/test.

\subsection{Dataset Characteristics}
\label{sec:chara}
Our {\it MultiSports} has several distinguishing characteristics compared with existing datasets.

\noindent {\bf Difficulty.} As discussed above, MultiSports is difficult in several aspects comparing to existing datasets: 1) multi-person situations of different concurrent actions, which prevents the model from distinguishing action categories only with backgrounds and requires models to capture subtly different motion cues; 2) a larger number of fine-grained categories with a long-tailed distribution; 3) the large variance of action instance duration, which makes it difficult to localize the temporal boundary; 4) the fast movement, large deformation and occlusion of actions in sports.

\noindent {\bf High Quality.} The videos of MultiSports are with high-resolution (720P or 1080P) competition records, which can preserve details of small humans and objects. Besides, with the help of our annotation team composed of professional athletes, our action categories and their corresponding action boundaries are precisely annotated. The professional annotators and careful quality control is able to provide consistent and clean annotations.

\noindent {\bf Diversity.} Our video clips are selected from competitions of different performance levels with diverse countries and genders, making the dataset less biased and good balanced for realistic sports analysis.

\noindent {\bf Application.} This task has many application scenarios for sports analysis. Combined with Re-ID techniques, we can automatically perform game commentary, AI referee and technical statistics. It can also assess the player abilities and provide information for developing the training plan and game strategy, and trading players between clubs.

\section{Experiments and Analysis}

\subsection{Datasets and Metrics}

\noindent\textbf{MultiSports benchmark.} To build a solid action detection benchmark, we manually split the instances into the training set, validation set, and testing set. Due to the long-tailed distribution of action instance numbers, following AVA~\cite{DBLP:conf/cvpr/GuSRVPLVTRSSM18}, we only evaluate on 60 classes that have at least 25 instances in validation and test splits to benchmark performance. We resize the whole dataset into 720P. In total, the current version contains 18,422 training instances from 1,574 clips and 6,577 validation instances from 555 clips. We provide the detailed ratio of training and validation instances for each sport in Appendix A. All those instances are selected from 3200 clips covering 247 competition records. Unless otherwise mentioned, we report the results trained on the training set and evaluated on the validation set. The testing set includes 1071 clips and we withhold the annotations in the public release.

\noindent\textbf{Metrics.} 
Following the standard practice~\cite{DBLP:conf/iccv/WeinzaepfelHS15,DBLP:conf/iccv/KalogeitonWFS17a}, we utilize frame-mAP and video-mAP to evaluate action detection performance. For video-mAP, we use the 3D IoU, which is defined as the temporal domain IoU of two tracks, multiplied by the average of the IoU between the overlapped frames. The threshold is 0.5 for frame-mAP, 0.2 and 0.5 for video-mAP.

\subsection{Spatio-temporal Action Detection Results}

\begin{table*}[h]
\footnotesize
  \begin{center}
   \resizebox{\textwidth}{!}{
  \begin{tabular}{c|c|ccc|ccc|ccc|c}
  \hline
 \toprule[1pt]
  \multirow{2}{*}{Method} & \multirow{2}{*}{Res} & \multicolumn{3}{c|}{MultiSports} & \multicolumn{3}{c|}{UCF101-24} &
  \multicolumn{3}{c|}{JHMDB} &
  \multicolumn{1}{c}{AVA}\\
  \cline{3-12}
  && F@0.5& V@0.2& V@0.5 & F@0.5& V@0.2& V@0.5& F@0.5& V@0.2& V@0.5& F-mAP@0.5\\
 \hline
 ROAD~\cite{DBLP:conf/iccv/SinghSSTC17} & 300 $\times$ 300 &3.90&0.00&0.00&70.7&69.8&40.9&-&60.8&59.7&-\\
 YOWO~\cite{DBLP:journals/corr/abs-1911-06644} & 224 $\times$ 224 &9.28&10.78&0.87&71.10&72.97&46.42&74.51&88.05&82.57&-\\
 MOC~\cite{DBLP:conf/eccv/LiW0W20} (K=7)  & 288 $\times$ 288 &22.51&12.13&0.77&78.0&82.8&53.8&70.8&77.3&77.2&-\\
 MOC~\cite{DBLP:conf/eccv/LiW0W20} (K=11)  & 288 $\times$ 288 &25.22&12.88&0.62&-&-&-&-&-&-&-\\
 \hline
 SlowOnly Det., $4\times16$~\cite{DBLP:conf/iccv/Feichtenhofer0M19} &short side 256 &16.70&15.71&5.50&-&-&-&-&-&-&20.02\\
 SlowFast Det., $4\times16$~\cite{DBLP:conf/iccv/Feichtenhofer0M19}& short side 256 &27.72&24.18&9.65&-&-&-&-&-&-&24.56\\
  \bottomrule[1pt]
  \end{tabular}
   }
   \vspace{1mm}
\caption{Comparison of the state-of-the-art methods on MultiSports, UCF101-24, JHMDB and AVA.}
\vspace{-4mm}
\label{table:sota}
  \end{center}
\end{table*}

\begin{figure*}[t]
\begin{center}
\includegraphics[width=16.5cm]{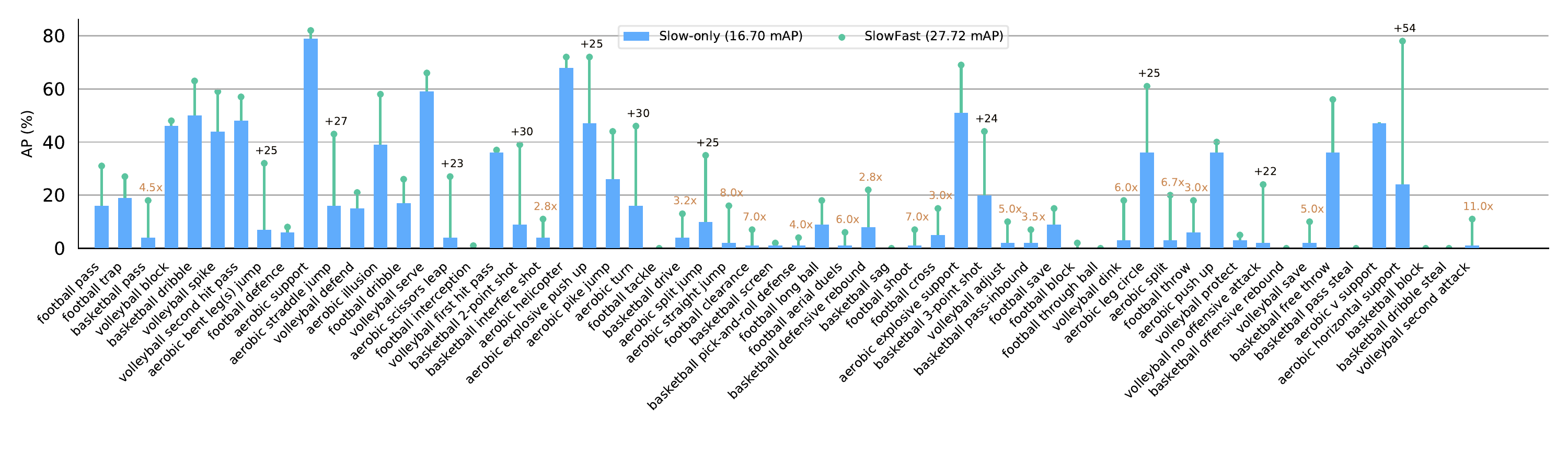}
\end{center}
\vspace{-6mm}
   \caption{SlowOnly vs. SlowFast frame-mAP. Categories are sorted by descending order on the number of instances.}
   \vspace{-2mm}
\label{fig:slowonly_slowfast_AP}
\end{figure*}
We evaluate several representative action detection methods on {\it MultiSports} and compare their performance on the UCF101-24~\cite{DBLP:journals/corr/abs-1212-0402}, JHMDB~\cite{DBLP:conf/iccv/JhuangGZSB13}, and AVA~\cite{DBLP:conf/cvpr/GuSRVPLVTRSSM18} in Table~\ref{table:sota}.  For SlowOnly Det. and SlowFast Det., we use the code in MMAction2~\cite{2020mmaction2}. We use the official released code for ROAD, YOWO and MOC. More details about the methods are provided in Appendix C. 
 
 For UCF101-24~\cite{DBLP:journals/corr/abs-1212-0402} and JHMDB~\cite{DBLP:conf/iccv/JhuangGZSB13}, which have dense annotations of high-level actions as MultiSports, we find that these methods achieve good performance on them but obtain low performance on MultiSports (frame-mAP of {\bf 25.22\%}, video-mAP@0.2 of {\bf 12.88\%} and video-mAP@0.5 of {\bf 0.62\%} for MOC~\cite{DBLP:conf/eccv/LiW0W20}). In our dataset, the largest performance drop occurs on ROAD~\cite{DBLP:conf/iccv/SinghSSTC17}, which is a frame-level action detector that performs action detection at each frame independently without exploiting temporal information.  UCF101-24~\cite{DBLP:journals/corr/abs-1212-0402} and JHMDB~\cite{DBLP:conf/iccv/JhuangGZSB13} have only one category per video. Characteristic visual scenes provide enough cues for predicting their coarse-grained actions. 
However, MultiSports has a similar background in the same sport, where the background fails to provide sufficient information for fine-grained action recognition. Meanwhile, our temporal boundary annotation is more precise and requires more accurate localization in temporal domain. 

For AVA~\cite{DBLP:conf/cvpr/GuSRVPLVTRSSM18}, which has only sparse annotations of atomic actions, we observe that the performance gap between SlowFast Det.~\cite{DBLP:conf/iccv/Feichtenhofer0M19} and SlowOnly Det.~\cite{DBLP:conf/iccv/Feichtenhofer0M19} on MultiSports is more evident than on AVA (frame-mAP gap of {\bf 11.02\%} vs. {\bf 4.54\%}). This indicates that the sports actions need a higher frame rate to capture fast motion at a finer temporal granularity. As shown in Figure~\ref{fig:slowonly_slowfast_AP}, many aerobic actions gain large absolute improvement, such as aerobic turn (+30 AP) and aerobic horizontal support (+54 AP). We analyze that aerobic actions' deformation and displacement is the largest among the four sports and benefit more from this finer temporal analysis. We also observe a large performance increase in other sports, such as basketball pass, football clearance and volleyball second attack, which have short temporal duration and intense motion. 

\subsection{Error Analysis}
\begin{figure}[t]
\begin{center}
\includegraphics[width=8.5cm]{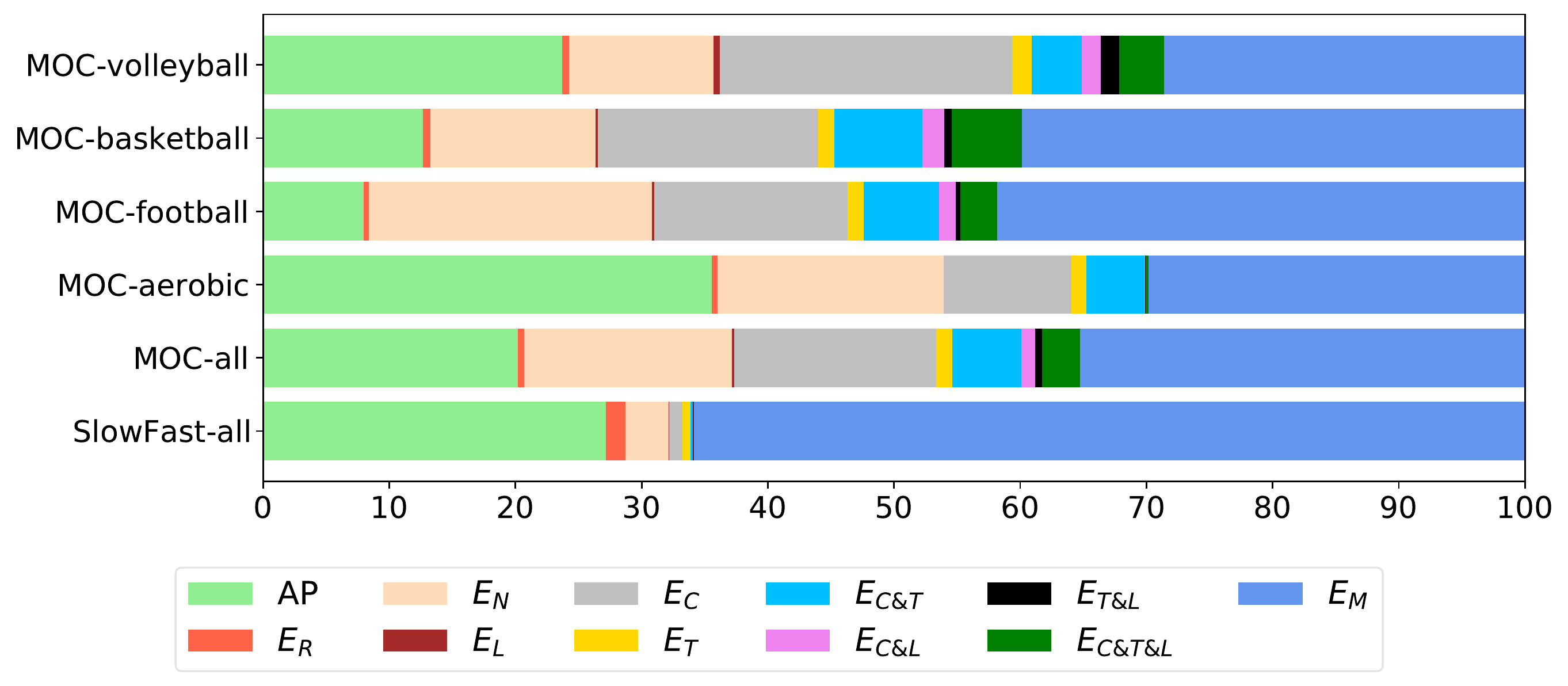}
\end{center}
\vspace{-1em}
   \caption{Error Analysis. AP is the correct detection. The threshold for a ground-truth matched by a detection is 0.1. Recall is $1-E_M$}
\label{fig:error_analysis}
\vspace{-2em}
\end{figure}

 In this section, we analyze the cause of errors to better understand {\it MultiSports}' challenges. Based on ACT~\cite{DBLP:conf/iccv/KalogeitonWFS17a} frame-mAP error analysis, which is designed for the dataset with one action category per video, we propose a new detailed error analysis in video-mAP. We classify the detection errors into 10 mutually exclusive categories to analyze which percentage of the mAP is lost. $\bf{E_R}$ : a detection result targets at a ground-truth tube that has already been matched. $\bf{E_N}$ : a detection result that has no spatial-temporal intersection with any ground-truth tubes and appears out of thin air. $\bf{E_L}$ : a detection result that has the correct action class, accurate temporal localization and inaccurate spatial localization. $\bf{E_C}$ : a detection result that has the wrong action class, accurate temporal localization and accurate spatial localization. $\bf{E_T}$ : a detection result that has the correct action class, accurate spatial localization and inaccurate temporal localization. $\bf{E_{C\&T}}$, $\bf{E_{C\&L}}$, $\bf{E_{T\&L}}$, $\bf{E_{C\&T\&L}}$: a detection that is inaccurate in corresponding aspects while acceptable in other aspect (if any). For example, $\bf{E_{C\&T}}$ refers to results with wrong action class, inaccurate temporal localization yet accurate spatial localization. $\bf{E_M}$ : ground-truth tubes that have not been matched by any detection results. The first nine categories cover the false positive predictions. The partition can be explained with a decision tree which is attached to our Appendix D. The code is provided at \href{https://github.com/MCG-NJU/MultiSports}{https://github.com/MCG-NJU/MultiSports}.
 
\begin{figure}[t]
\begin{center}
\includegraphics[width=8.5cm]{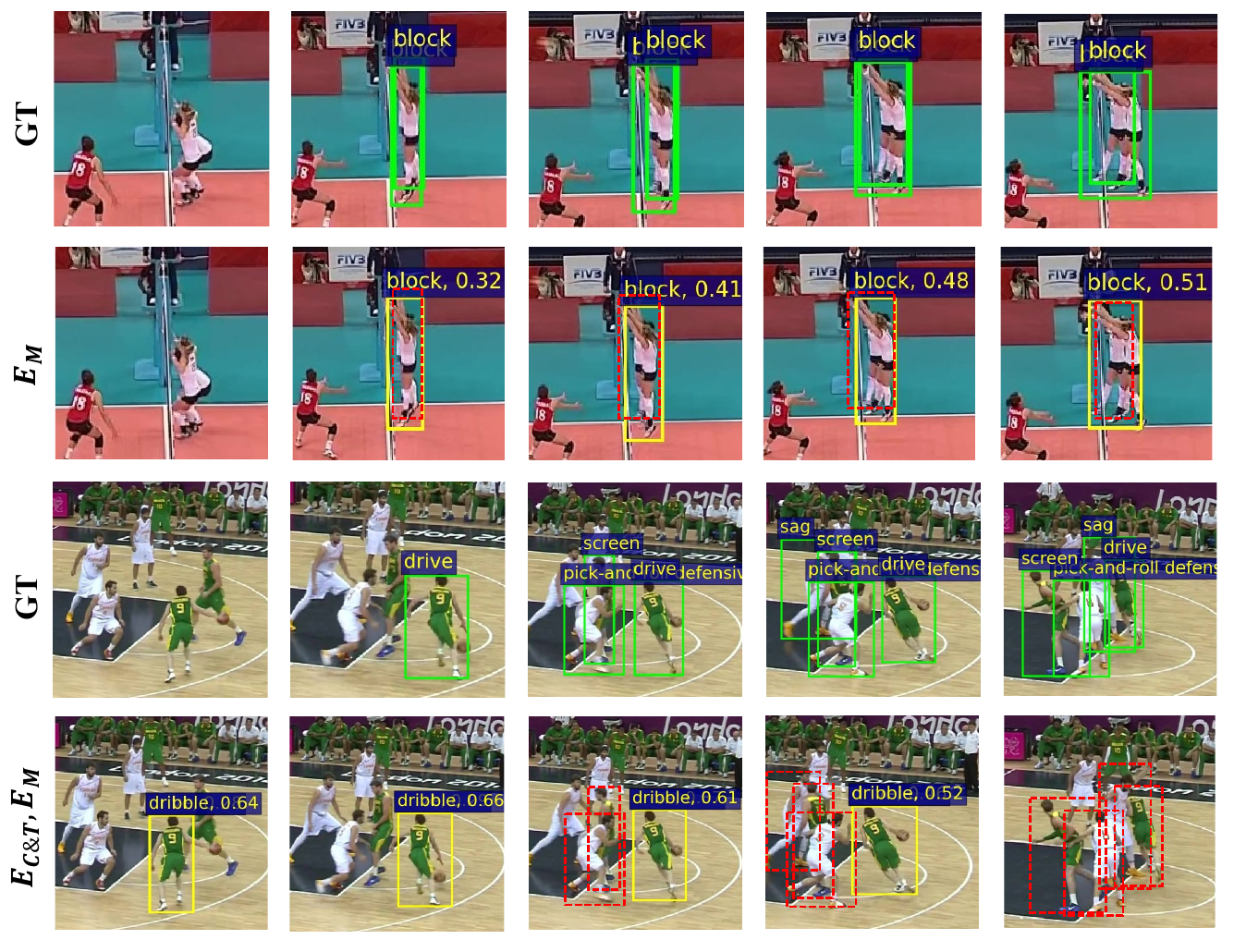}
\end{center}
\vspace{-1em}
   \caption{Visualization of typical errors in MultiSports. Green boxes are the ground-truths. Yellow boxes are the detections. Red boxes are the missed ground-truths. 1st and 2nd row: missed detection due to occlusion. 3rd and 4th row: $E_{C\&T}$: drive is misclassified as dribble and also has inaccurate action boundary; $E_M$: missed detections of screen, pick-and-roll defensive and sag.}
\label{fig:error_visualization}
\vspace{-1em}
\end{figure}

 As shown in Figure~\ref{fig:error_analysis}, despite the relatively low recall, SlowFast Det. achieves higher video-mAP than MOC because it makes much fewer false positive predictions. This can be explained by the fact that SlowFast Det. uses Faster RCNN~\cite{DBLP:journals/pami/RenHG017} finetuned on MultiSports as person detector to greatly avoid the person boxes without actions. However, there are still many hard examples missed by SlowFast Det. For MOC, $E_C$ and $E_N$ are the most common errors among false positive detection results, indicating the difficulty of our fine-grained action classification. Detection results with $E_N$ errors means the model indeed detects the person spatio-temporally but unable to identify the action class correctly as the background class. $E_N$ error is also a result of the training strategy of MOC where only the frames temporally inside action instances are sampled for training, so that although there are negative samples in other spatial location of these frames, the detector does not have enough amount of negative samples for people without doing any sports action. What is more, $E_{C\&T}$, $E_{C\&T\&L}$ and $E_{T}$ are also a large portion of the rest errors (where $E_{C\&T} > E_{C\&T\&L} > E_{T}$), indicating more temporal errors with inaccurate action boundaries than spatial localization errors for current methods. Therefore we need a more effective way of modeling temporal boundary. Typical error visualization is shown in Figure~\ref{fig:error_visualization}.

\subsection{Ablation Study}

\noindent\textbf{How important is temporal information?}
The tubelet length $K$ is important in MOC~\cite{DBLP:conf/eccv/LiW0W20} and we report results on UCF101-24~\cite{DBLP:journals/corr/abs-1212-0402} and {\it MultiSports} with different $K$ in Table~\ref{table:temporal}. For frame-mAP, we can find that MultiSports can benefit more from longer temporal context than UCF101-24, in spite of the shorter action duration of MultiSports than UCF101-24 as shown in Table~\ref{tab:stat}. For video-mAP, the result does not increase as frame-mAP. We analyze there are two reasons. First, predicting movement in MOC is harder with longer input length. What is worse, the categories in MultiSports have large deformation and displacement, and MOC Movement Branch can not predict them accurately, which harms the video level detection seriously. Second, Figure~\ref{fig:interval} shows the variability of action duration. The ratio is $9\%$ for instances duration less than 7 and $23\%$ for less than 11. The fixed clip length $K$ (e.g. 11) will damage temporal detection ability. So, we need to consider longer temporal context, more accurate movement estimation and flexible temporal detection for MultiSports.

\noindent\textbf{Which action categories are challenging?}
Figure~\ref{fig:slowonly_slowfast_AP} shows that not all categories yield better performance with more training samples. Categories highly correlated with scenes (such as basketball free throw) or aerobics basic categories (such as aerobic horizontal support and V support) can still achieve high performance with fewer samples. Note that aerobics contains basic and complex categories, where complex action combines the motion of basic action and its own core motion, thus longer temporal context is required for these complex actions. In contrast, categories with short temporal duration and intense motion (such as football pass, basketball pass and football interception) achieve low performance even though with lots of training samples. By observing the confusion matrix in Appendix D, we summarize other common challenges: (1) Context modeling, such as basketball 2-point shot vs. 3-point shot (2) Reasoning, such as for volleyball protect vs. defend, we need to focus on whether the ball was blocked back or was spiked by an opponent several frames earlier. (3) Long temporal modeling, such as football long ball vs. pass, they have the similar motion but need to identify how long the ball will be passed. 

\begin{table}[t]
\footnotesize
  \begin{center}
  \begin{tabular}{c|ccc|ccc}
  \hline
 \toprule[1pt]
  \multirow{2}{*}{K} & \multicolumn{3}{c|}{MultiSports} & 
  \multicolumn{3}{c}{UCF101-24}\\
  \cline{2-7}
  & F@0.5& V@0.2& V@0.5& F@0.5&V@0.2&V@0.5\\
 \hline
 1&14.61&12.53&1.06&68.33&65.47&31.50\\
 3&17.22&11.88&0.76&69.94&75.83&45.94\\
 5&19.29&11.81&{\bf 0.98}&71.63&77.74&49.55\\
 7&22.51&12.13&0.77&{\bf 73.14}&{\bf 78.81}&{\bf 51.02}\\
 9&24.22&11.72&0.57&72.17&77.94&50.16\\
 11&\bf{ 25.22}&{\bf 12.88}&0.62&-&-&-\\
 13&24.28&11.23&0.57&-&-&-\\
  \bottomrule[1pt]
  \end{tabular}
  \vspace{1mm}
  \caption{Exploration study of MOC on the MultiSports and UCF101-24 with different tubelet length K. }
  \label{table:temporal}
  \vspace{-6mm}
  \end{center}
\end{table}
\begin{table}[t]
\footnotesize
  \begin{center}
  \begin{tabular}{c|ccc|c}
  \hline
 \toprule[1pt]
  \multirow{2}{*}{Estimation} & \multicolumn{3}{c|}{MultiSports} & 
  \multicolumn{1}{c}{AVA}\\
  \cline{2-5}
  & F@0.5& V@0.2& V@0.5& F-mAP@0.5\\
 \hline
 Untrimmed&27.72&24.18&9.65&22.57\\
 Trimmed&38.71&24.95&18.34&24.56\\
  \bottomrule[1pt]
  \end{tabular}
  \vspace{1mm}
  \caption{Test SlowFast Det. on AVA and MultiSports with trimmed way and untrimmed way.}
  \vspace{-6mm}
  \label{table:test}
  \end{center}
\end{table}
\noindent\textbf{Trimmed vs. untrimmed settings.}
{\it MultiSports} has well-defined and high-quality temporal boundaries. We evaluate the performance of SlowFast Det. under both the untrimmed and trimmed setting on MultiSports and AVA datasets.  The results are reported in Table~\ref{table:test}. The trimmed setting only evaluates the performance on the frames having annotations and the untrimmed setting reports the performance on all frames. We find that it only drops $2\%$ on AVA while $11\%$ on our dataset, which indicates that temporal localization is really important in our dataset. In addition, video-mAP@0.5 drops far more than video-mAP@0.2. This demonstrates that temporal localization is important for high-quality action tube detection. 

\section{Conclusion}

In this paper, we have introduced the {\it MultiSports} dataset with dense spatio-temporal annotations of actions from four sports. MultiSports distinguishes from the existing action detection datasets in many aspects: 1)  raising new challenges for recognizing fine-grained action classes; 2) requirement of accurate localization of well-defined boundaries in multiple-person situations; 3) high quality video data and dense annotations; 4) potential applications in sports analysis; 5) less biased dataset with high diversity in competition levels, countries and genders. We have empirically investigated several action detection baseline methods on the MultiSports dataset. Our error analysis and ablation studies on the detection results uncover several insightful findings that are beneficial for the future research of spatio-temporal action detection.

\paragraph {\bf Acknowledgements.} This work is supported by National Natural Science Foundation of China (No. 62076119, No. 61921006), Program for Innovative Talents and Entrepreneur in Jiangsu Province, and Collaborative Innovation Center of Novel Software Technology and Industrialization. Thanks to professional athletes of Nanjing University varsities and MCG students for annotating this dataset.

\section*{Appendix A: More Dataset Details}

\subsection*{A.1 Train split vs. Validation split}

\begin{table*}[htbp]
    \footnotesize
    \centering
    \scalebox{1.4}{
    \begin{tabular}{c|ccccc}
    \toprule[0.75pt]
     & Volleyball & Football & Basketball& Aerobic& All\\
     \hline
    instance ratio&3549:1294&6144:2153&4532:1715&4197:1415&18422:6577 \\
    clip ratio&402:130&402:132&379:147&391:146&1574:555 \\
    competition ratio&32:11&36:12&34:14&23:8&125:45 \\
    \bottomrule[0.75pt]
    \end{tabular}
    }
    \vspace{4mm}
    \caption{Train split vs Validation split}
    \label{tab:class_id}
\end{table*}
In order to guarantee enough instances for each class despite the severely unbalanced distribution, we artificially split the instances into the training set and the validation set in Table~\ref{tab:class_id}. To avoid data leakage from the training set to the validation/testing set, we ensure that data from the same match should be used for only one purpose. In other words, clips in the validation set cannot come from the matches covered in the training set. Unless otherwise mentioned, we report the results trained on the training set and evaluated on the validation set.

\subsection*{A.2 Comparison with other type of Dataset}

MEVA~\cite{DBLP:conf/wacv/CoronaOCH21} is a new security dataset, whose data is from RGB and thermal IR cameras, UAV footage and GPS locations for the actors. It defines 37 activities (66 for {\it MultiSports}) with 17055 instances (37701 for {\it MultiSports}), where 29 activities are about person and 8 activities are about vehicle. The categories in this dataset are atomic, such as $person\_close\_trunk$ and $person\_stand\_up$, which are different from our fine-grained and complicated sports categories. What's more, most of the categories in MEVA are daily actions, whose deformation and displacement are not large. Although it is a multi-person dataset, we believe our {\it MultiSports} can bring new challenges different from MEVA.

\section*{Appendix B: More Ablation Study}
\noindent\textbf{How the well-defined and high quality temporal boundary help?}
We add some temporal noise to the train set GT. For a $L$-frame length instance, we randomly choose a new length new\_L from (1, L) and then the start point offset from (0, L-new\_L). We sample the new annotation from the original. Other settings are kept the same. From the Table~\ref{tab:ablation}, we find the performance is much worse without well-defined temporal boundaries. It can conclude that our {\it MultiSports} has well-defined and high quality temporal annotations, which can help improve the performance and promote the algorithms to localize the boundary more accurately. 

\begin{table}[h]
\footnotesize
  \begin{center}
   \resizebox{0.4\textwidth}{!}{
  \begin{tabular}{c|c|ccc}
  \hline
 \toprule[1pt]
  \multirow{2}{*}{Method} & \multirow{2}{*}{GT Noise} & \multicolumn{3}{c}{MultiSports} \\
  \cline{3-5}
  && F@0.5& V@0.2& V@0.5 \\
 \hline
 MOC (K=7)  & \CheckmarkBold  &13.71&8.59&0.63\\
 MOC (K=7)  &\XSolidBrush &22.51&12.13&0.77\\
 \hline
 SlowOnly Det., $4\times16$ & \CheckmarkBold &12.60&8.98&3.05\\
 SlowOnly Det., $4\times16$ &\XSolidBrush&16.70&15.71&5.50\\
  \bottomrule[1pt]
  \end{tabular}
   }
  \end{center}
  \caption{Exploration on the effect of the temporal boundary noise.}
  \label{tab:ablation}
\end{table}

\section*{Appendix C: Method Details}

\noindent{\textbf{ROAD}}~\cite{DBLP:conf/iccv/SinghSSTC17} is a deep-learning framework for real-time action localisation and classification. It adopts SSD~\cite{DBLP:conf/eccv/LiuAESRFB16} to regress and classify action detection boxes in each frame independently, which does not utilize temporal information. Then, the frame detections are linked into action tubes by an online algorithm. Here we use the python linking code provided by MOC~\cite{DBLP:conf/eccv/LiW0W20} instead of the original MATLAB code. Following the settings of ROAD on UCF101-24~\cite{DBLP:journals/corr/abs-1212-0402}, we use an ImageNet pre-trained VGG16~\cite{DBLP:journals/corr/SimonyanZ14a} network. We first try an initial learning rate of 1e-4 as their setting on UCF101-24, but the loss diverges into infinity after 20 iterations. The reported experiment on our {\it MultiSports} adopts an initial learning rate of 1e-5. We use SGD optimizer and the learning rate is reduced to its $\frac{1}{10}$ after 30000, 60000 iterations, which is the same as their practice on UCF101-24. The maximum iteration number is 120000.

\noindent{\textbf{YOWO}}~\cite{DBLP:journals/corr/abs-1911-06644} is a frame-level action detector with two branches. A 2D-CNN branch extracts the spatial features of the key frame while a 3D-CNN branch extracts spatio-temporal features of the key frame and the previous n (n=16) frames. Then, the features of two branches are fused by a channel fusion and attention mechanism(CFAM) module and finally passed to a convolution layer to predict the action class and bounding box in Yolov2~\cite{DBLP:conf/cvpr/RedmonF17} manner. Finally, the frame detections are linked into action tubes by a dynamic programming algorithm. Note that the linking algorithm in YOWO is trimmed, thus we use the same linking algorithm as MOC on {\it MultiSports}. We use 2D Darknet-19 backbone pretrained on PASCAL VOC~\cite{DBLP:journals/ijcv/EveringhamEGWWZ15} and 3D ResNeXt-101 backbone pre-trained on Kinetics~\cite{DBLP:conf/cvpr/CarreiraZ17}. To utilize multiple GPUs, we modified the batch size to 80 and the initial learning rate to 8e-4. Following the training strategy of YOWO on UCF101-24~\cite{DBLP:journals/corr/abs-1212-0402}, we adopt SGD optimizer and the learning rate is reduced to its $\frac{1}{2}$ after 30000, 40000, 50000, 60000 iterations. The epoch maximum is set to 5. Note that YOWO only estimates performance on the frames having annotations, thus frame-mAP we report on UCF101-24 is much lower than in the original paper.

\noindent{\textbf{MOC}}~\cite{DBLP:conf/eccv/LiW0W20} is an anchor-free tubelet-level action detector with three branches, which firstly takes $K$ frames as input, then outputs $K$ frame tubelet results and finally links these tubelets into tubes with a common matching strategy. We use DLA34~\cite{DBLP:conf/cvpr/YuWSD18} as the backbone network, which is pre-trained on COCO~\cite{DBLP:conf/eccv/LinMBHPRDZ14}. Following the training strategy of MOC on UCF101-24~\cite{DBLP:journals/corr/abs-1212-0402}, we use the Adam optimizer with the learning rate 5e-4. The learning rate is reduced to its $\frac{1}{10}$ after epoch 6 and 8. The epoch maximum is set to 12.

\noindent{\textbf{SlowFast Det.}}~\cite{DBLP:conf/iccv/Feichtenhofer0M19} firstly uses a person detector on the key frame to localize for region proposal.  Then, each 2D RoI at the key frame is extended into a 3D RoI by replicating it along the temporal dimension. Finally, it extracts RoI features from the backbone features for predicting category. The person detector is a Faster R-CNN with a ResNeXt-101-FPN~\cite{DBLP:conf/cvpr/XieGDTH17, DBLP:conf/cvpr/LinDGHHB17} backbone, which is pre-trained on ImageNet~\cite{DBLP:conf/cvpr/DengDSLL009} and the COCO human keypoint images~\cite{DBLP:conf/eccv/LinMBHPRDZ14}. The backbone is the variant of  SlowFast or SlowOnly, which sets the spatial stride of $res_5$ to 1 and uses a dilation of 2 for its filters. Note that we use the code in MMAction2~\cite{2020mmaction2}. The results on AVA~\cite{DBLP:conf/cvpr/GuSRVPLVTRSSM18} and our {\it MultiSports} in the paper are all produced by it. We use the pre-computed proposals for AVA from previous work~\cite{DBLP:conf/iccv/Feichtenhofer0M19,DBLP:conf/cvpr/WuF0HKG19}. Following previous work~\cite{DBLP:conf/iccv/Feichtenhofer0M19,DBLP:conf/cvpr/WuF0HKG19}, we fine-tune the person detector on our {\it MultiSports} with MMDetection~\cite{DBLP:journals/corr/abs-1906-07155}. We use the SGD optimizer with the learning rate 0.0025 and finetune 2 epochs on our {\it MultiSports}. The person detector produces 96.16 AR@100 on our {\it MultiSports} validation set. The detected boxes with confidence of $>$ 0.9 are selected for action detection on both datasets. Our backbones are based on ResNet50, which are pre-trained on Kinetics-400~\cite{DBLP:conf/cvpr/CarreiraZ17}. The $T \times \tau$ is set to $4 \times 16$. The $\alpha$ is set to 8 for SlowFast. We use a step-wise learning rate, reducing the learning rate $10 \times$ after epoch 6 and 7. We train for 8 epochs with a linear warm-up for the first 5 epochs, where the result is similar with that of training 20 epochs and a lot of training time is saved. The initial learning rate is set to 0.1125 for SlowFast and 0.2 for SlowOnly. SlowFast and Slowonly Det. use the same link algorithm as MOC.

\section*{Appendix D: Error Analysis}
\begin{figure*}[t]
\begin{center}
\includegraphics[width=18.5cm]{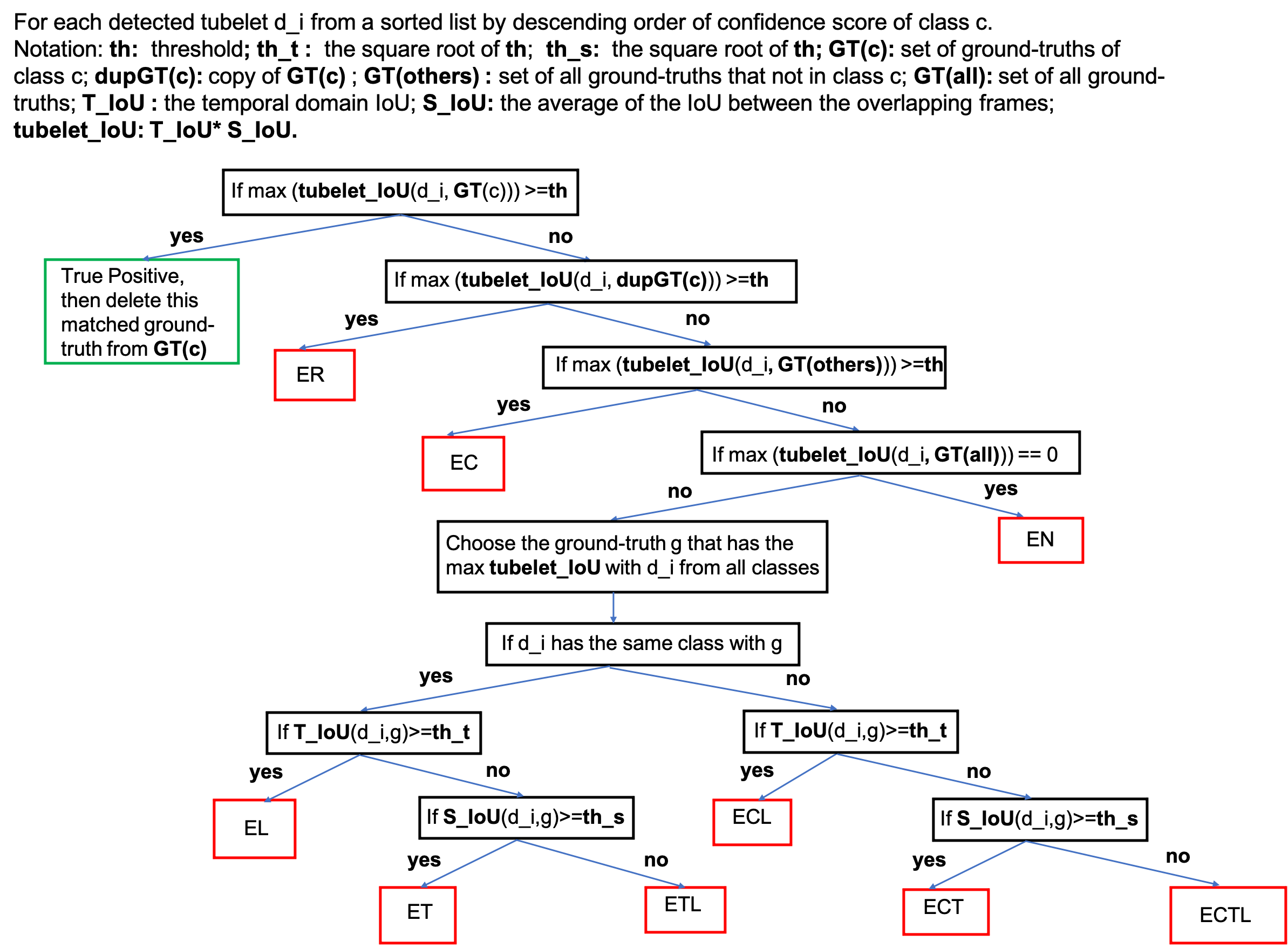}
\end{center}
\vspace{-1em}
   \caption{Error Tree}
\label{fig:error-tree}
\vspace{-1em}
\end{figure*}

\begin{figure*}[t]
\begin{center}
\includegraphics[width=17.5cm]{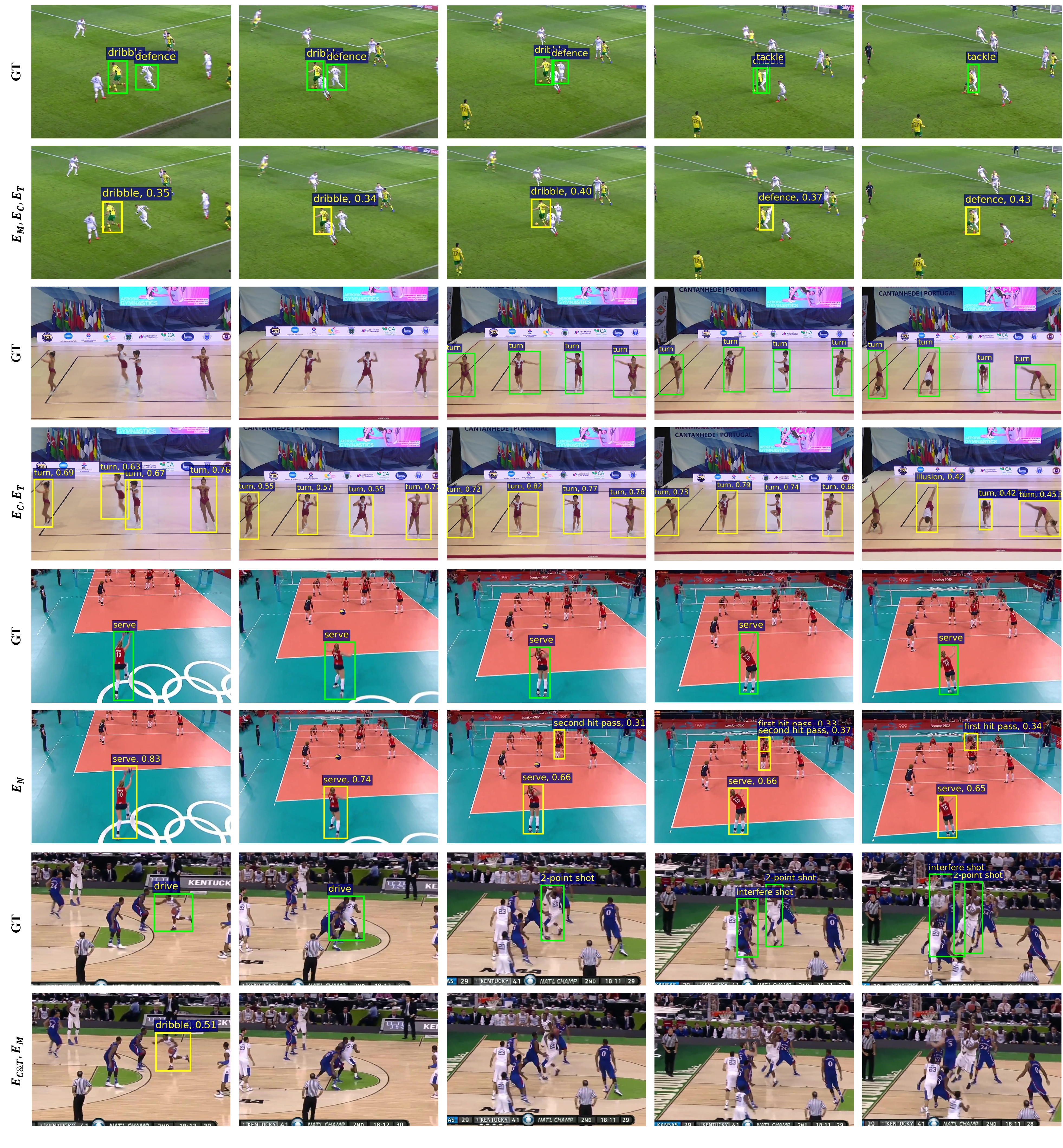}
\end{center}
\vspace{-1em}
   \caption{More detailed visualizations on our {\it MultiSports} dataset with our novel error categories of video-mAP. Green boxes are the ground-truths. Yellow boxes are the detections. 1st and 2nd row: $E_M$: missed detection of defence; $E_C$: tackle is misclassified as defence; $E_T$: dribble has inaccurate action ending boundary. 3rd and 4th row: $E_C$: turn is misclassified as illusion in the last picture in 4th row; $E_T$: turn has inaccurate action boundary. 5rd and 6th row: $E_N$: detection results contain that athletes actually doing none of sports actions but the model identifies first hit pass and second hit pass for them. 7rd and 8th row: $E_{C\&T}$: drive is misclassified as dribble and also has inaccurate action boundary; $E_M$: missed detections of interfere shot and 2-point shot.}
\label{fig:visualization}
\vspace{-1em}
\end{figure*}

\subsection*{D.1 Error Tree}
 To further understand the difficulty in our {\it MultiSports} dataset, we classify the detection errors into 10 different categories in a tree structure as shown in Figure~\ref{fig:error-tree} (code in $VideomAP\_error.py$), which are:
\begin{itemize}
	\item $E_R$ (Errors of repeated detections): a detection result that has tubelet IoU larger than a threshold and the right action class with some ground-truth tubelets, but the ground-truths have been matched by other detection results before with a confidence score larger than it.
	\item $E_N$ (Errors of not matched): a detection result that has no intersection with any ground-truth tubelets of any class, indicating there should be no detection results but it appears out of thin air.
	\item $E_L$ (Errors of spatial localization): a detection result that has the same action class and temporal IoU larger than a threshold with some ground-truth, but it has a low average spatial bounding box IoU in the area of the temporal intersection of ground-truth tubelets and it so that a lower tubelet IoU than the required threshold.
	\item $E_C$ (Errors of classification): a detection result that has the tubelet IoU larger than a threshold with a ground-truth, but its action class is not the same with the ground-truth's class.
	\item $E_T$ (Errors of temporal localization): a detection result that has the same action class and average spatial bounding box IoU larger than a threshold with some ground-truth in the area of the temporal intersection of ground-truth tubelets and it, but low temporal IoU with ground-truths so that a lower tubelet IoU than the required threshold.
	\item $E_{C\&T}$ (Errors of both classification and temporal localization): a detection result that has average spatial bounding box IoU larger than a threshold with some ground-truth tubelets in the area of the temporal intersection of ground-truth tubelets and it, but both low temporal IoU and wrong action class.
	\item $E_{C\&L}$ (Errors of both classification and spatial localization): a detection result that has temporal IoU larger than a threshold with some ground-truth tubelets, but both wrong action class and low average spatial bounding box IoU with some ground-truth in the area of the temporal intersection of ground-truth tubelets and it.
	\item $E_{T\&L}$ (Errors of both temporal and spatial localization): a detection result in which we first select the ground-truth tubelet from all action classes that has the maximum tubelet IoU with the detection result, then we find they share the same action class, but both temporal IoU and average IoU of spatial bounding boxes lower than a threshold.
	\item $E_{C\&T\&L}$ (Errors of classification, temporal and spatial localization): a detection result that has some intersection with some ground-truth tubelets, which is different with EN, but wrong action class and both the temporal and average bounding box IoU lower than a threshold.
	\item $E_M$ (Errors of missed detections): ground truth tubelets that have not been matched by any detection results.
\end{itemize}
\subsection*{D.2 More Visualization of Error Analysis}
As shown in Figure~\ref{fig:visualization}, we collect more visualizations of MOC(K=11) as a supplementary of Figure 7 in our paper. 

\subsection*{D.3 Confusion Matrix}
\begin{figure*}[ht]
		\begin{center}
		\subfigure[Aerobic]{
				\begin{minipage}{8.5cm}
					\centering
					\includegraphics[width=8.5cm]{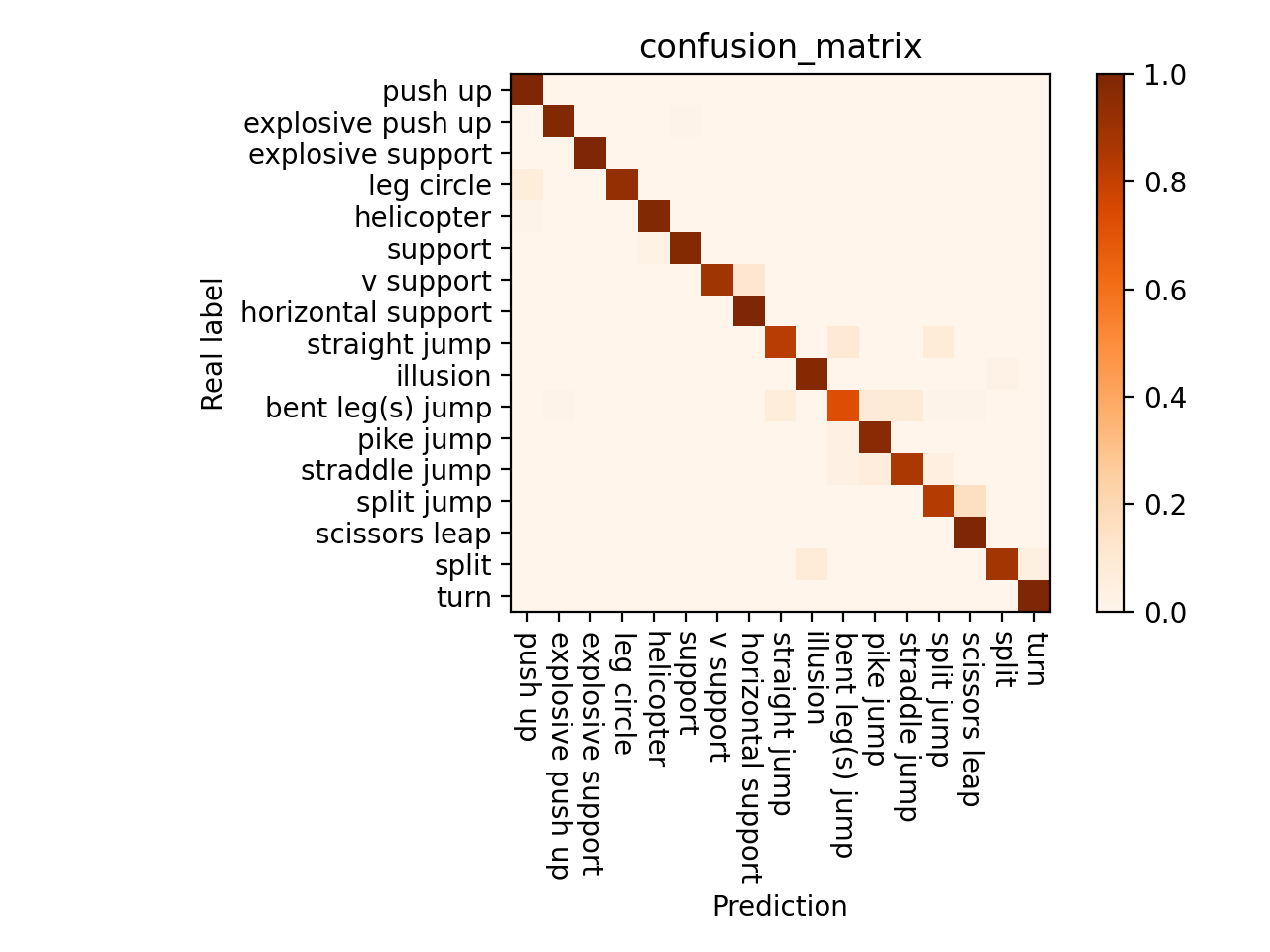}
				\end{minipage}
		}
		\subfigure[Volleyball]{
			\begin{minipage}{8.5cm}
				\centering
				\includegraphics[width=8.5cm]{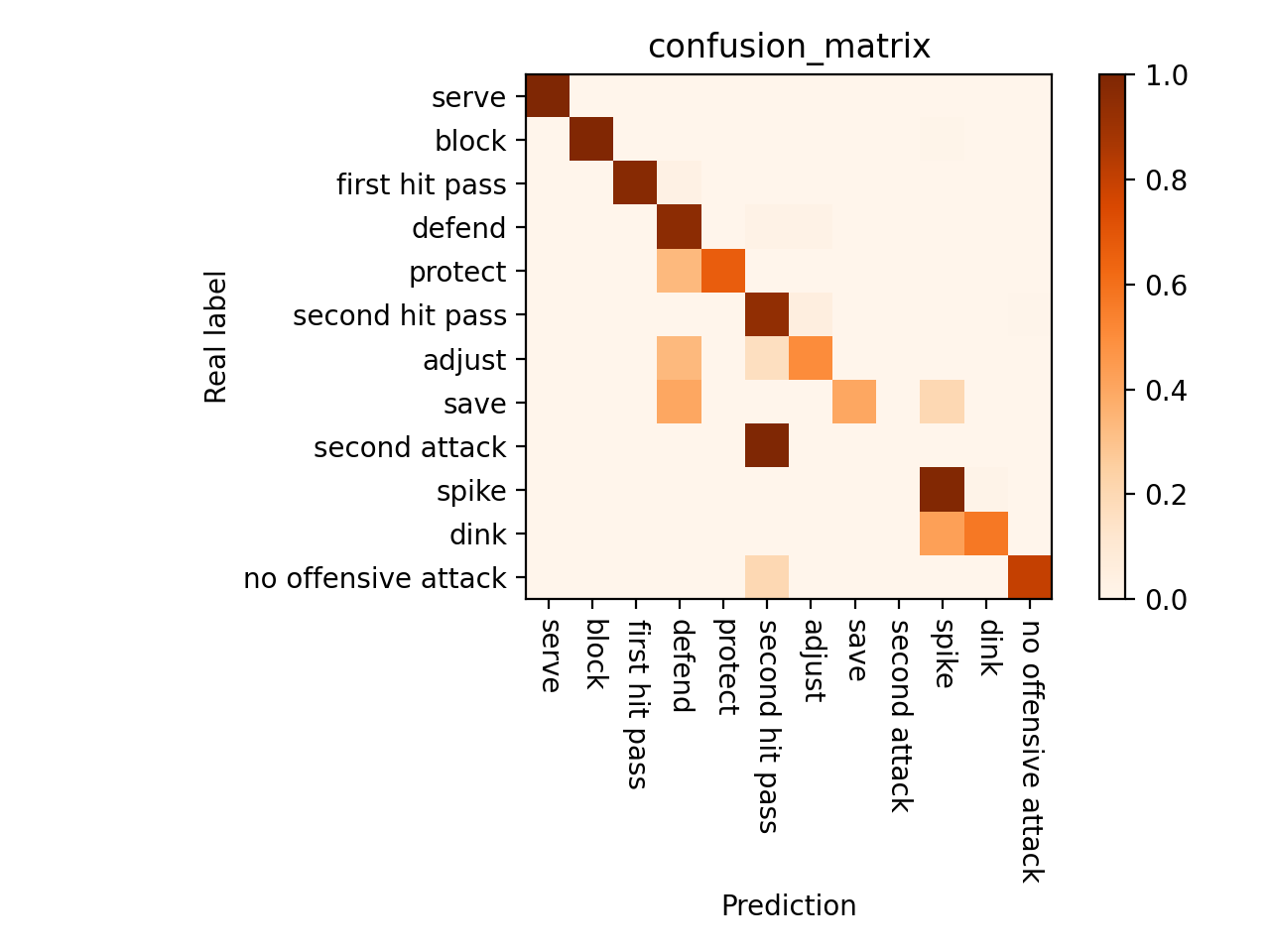}
			\end{minipage}
		}
		\hspace{1in}
		\subfigure[Basketball]{
			\begin{minipage}{8.5cm}
				\centering
				\includegraphics[width=8.5cm]{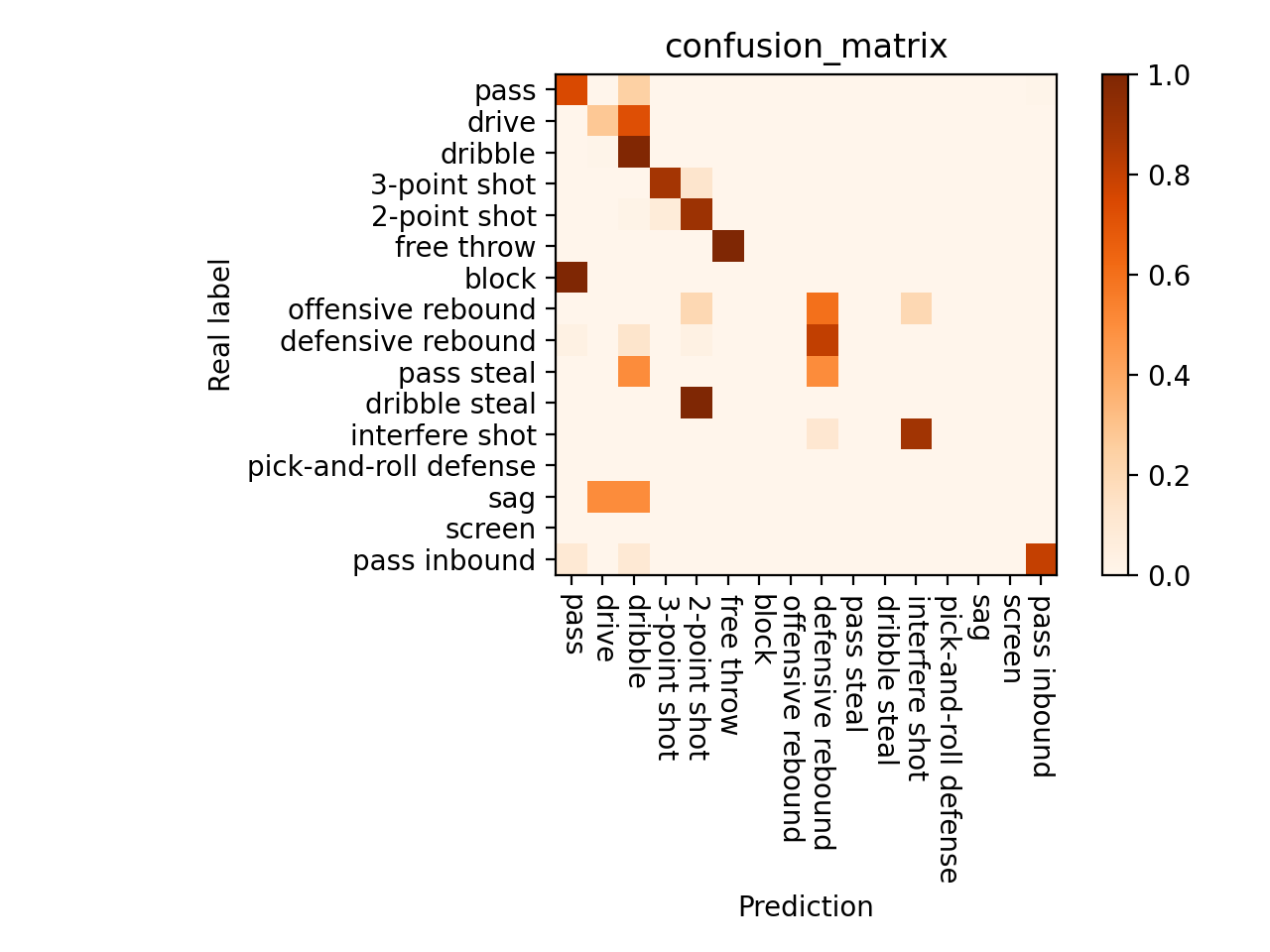}
			\end{minipage}
		}
		\subfigure[Football]{
			\begin{minipage}{8.5cm}
				\centering
				\includegraphics[width=8.5cm]{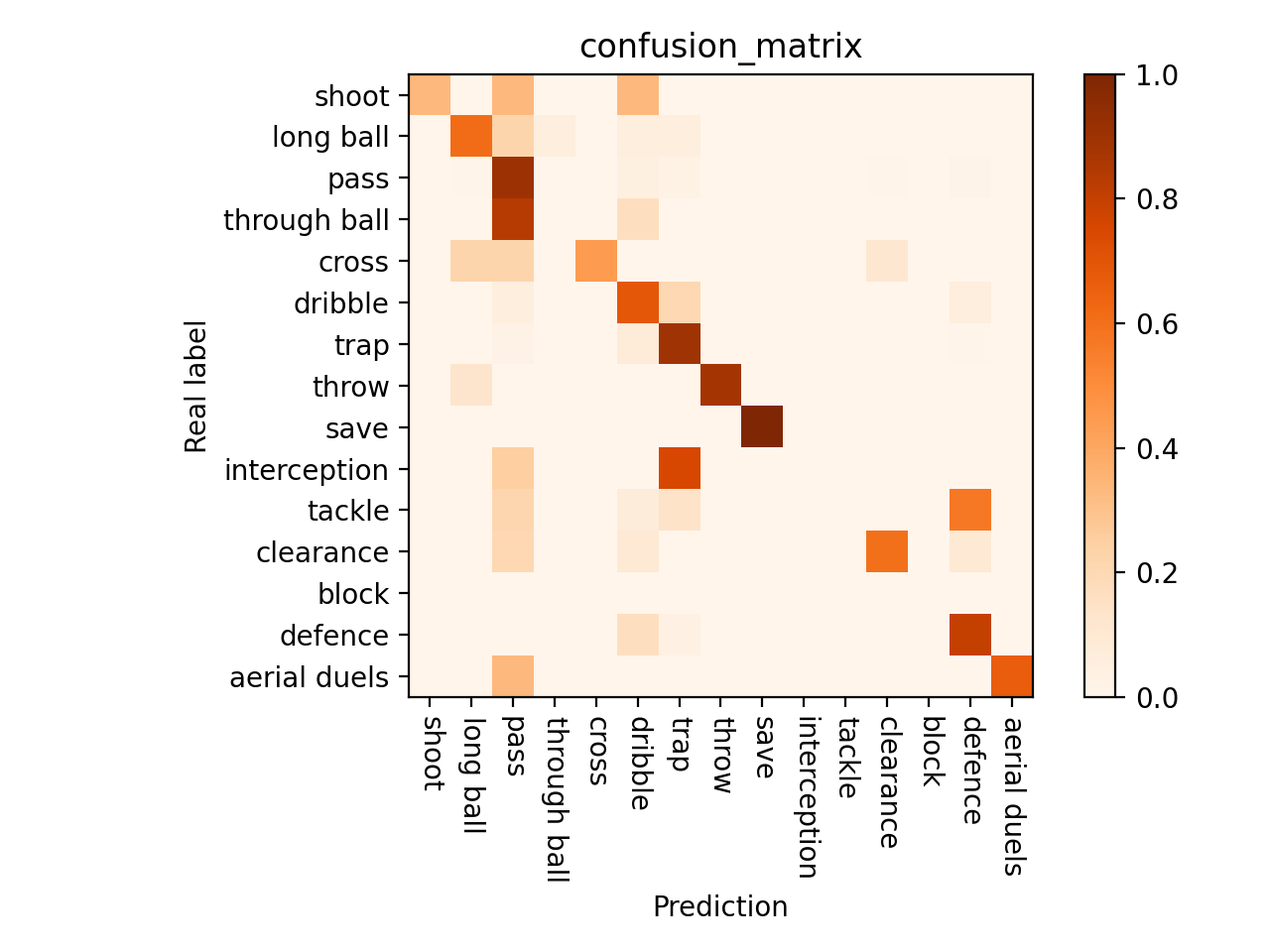}
			\end{minipage}
		}
		\end{center}
		\vspace{-1em}
		   \caption{Confusion Matrix of SlowFast Det. on different sports.}
\label{fig:confusion}
\vspace{-1em}
\end{figure*}
We draw the confusion matrices of the predictions which are classified into $AP$ and $E_C$ in Figure~\ref{fig:confusion}.  We observe that the aerobic performs best because its categories relate only to individual actors. Actions having similar motions but different spatio-temporal contexts tend to confuse. For example, 1) drive vs. dribble in basketball, drive emphasizes on breaking through defender and being closer to the basket, which needs to model person-person interaction and spatial localization; 2) through ball vs. pass in football, through ball will break through the opponent's line of defense and be passed in front of the teammate, which needs long-term temporal modeling and reasoning. 3) offensive rebound vs. defensive rebound, the difference is whether the offensive player or defensive player gains control of the ball; 4) defend vs. protect in volleyball, we need to focus on whether the ball was blocked back or was spiked by an opponent several frames earlier.

\section*{Appendix E: Annotation Documentation}
\subsection*{E.1 Aerobic Gymnastics}
\begin{figure*}[t]
\begin{center}
\includegraphics[width=17.5cm]{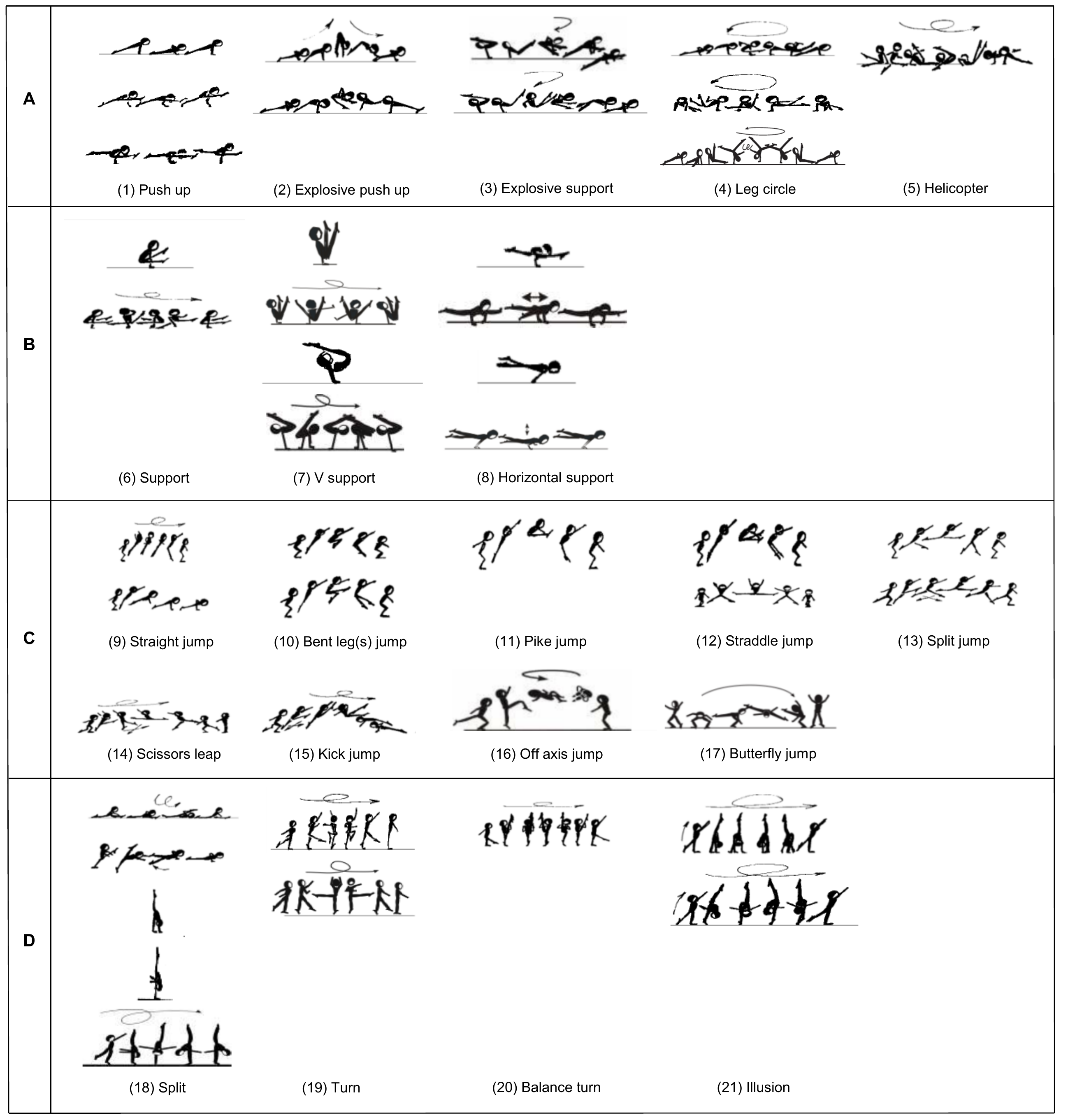}
\end{center}
\vspace{-1em}
   \caption{Diagrams of each difficulty element in aerobic gymnastics.}
\label{fig:aerobic_demo}
\vspace{-1em}
\end{figure*}

There are four groups of difficulty elements in aerobic gymnastics, namely dynamic strength, static strength, jumps \& leaps, and balance \& flexibility.  We pick out 21 elements to form the aerobic categories of our {\it MultiSports}. The following is a detailed definition of these categories, a simplified version of the definition in \cite{de2017aerobic}.

{\bf Group A: Dynamic Strength.} All elements in Group A ending in a split position, must have both hands on each side of the body on the floor. 
\begin{itemize}
    \item {\bf Push up}: Starting and/or finishing: one or both hands are in contact with the floor, elbows extended. Shoulders must be parallel to the floor; head in line with the spine and pelvis tucked with abdominal muscles contracted. Flexion of elbows: All push-ups must have, at the end of the downwards phase, a maximum distance of 10cm from the chest to the floor. The downward and/or the upward phase of a push up must be controlled with shoulders parallel to the floor. Lateral and Hinge push up, 4 phases have to be shown. Wenson push up: one leg on the upper part of the arm (Triceps) of the same side.
    \item {\bf Explosive push up}: 1) A Frame: Pike position in the airborne phase (60° between trunk and legs). 2) Cut: While airborne, the legs straddle sideways and forward to land extended in rear support, feet lifted off the floor during the skill.
    \item {\bf Explosive support}: Back support on the floor, back parallel to the floor, extending the legs upward and forward with a flight phase. Impulse from High V support position, airborne phase and landing to push up or split position.
    \item {\bf Leg circle}: The starting position must be from free front support on both hands; the hips must be lifted and extended during the full rotation. Feet may not touch the floor before the completion of the circle. 1) Leg circle: the hips must be lifted and extended. 2) Flair: legs straddle, the hips must be lifted and extended during the full rotation. Feet may not touch the floor before the completion of the circle.
    \item {\bf Helicopter}: After alternative leg circles, legs close to the chest, body alignment on the upper back (feet off the floor). The legs are extended upward and forward. ½ twist initiated from the feet is made to land in push up or wenson or split.
\end{itemize}

{\bf Group B: Static Strength.} These elements demonstrate isometric strength and must be held for 2 seconds. In the case of turns in support, the support must be held for 2 seconds either at the start, during or end of the turn. The body is fully supported by one or both arms and only the hands are in contact with the floor. Feet and/or hips must not touch the floor during the whole skill. While in support, the hands must be flat on the floor.
\begin{itemize}
    \item {\bf Support}: 1) Straddle support: Legs must be straight parallel to the floor in Straddle position (90°minimum). 2) L support: Legs must be straight together and parallel to the floor. 
    \item {\bf V support}: 1) Straddle V support: Hips are flexed and legs straddled 90° open and vertical, minimum width 90°. 2) V support: Hips are flexed and legs are together vertical. 3) High V support:  The back is parallel to the floor.
    \item {\bf Horizontal support}: 1) Wenson support: the body is extended parallel to the floor, one leg supported on the upper part of the Triceps. 2) Planche: the body is supported on both hands with straight arms, not more than 20° above parallel.
\end{itemize}

{\bf Group C: Jumps \& Leaps.} All jumps and leaps must demonstrate explosive power and maximum amplitude. All jumps that can be performed from 1 foot or two feet will be considered as the same element and will receive the same value.  This applies also for landing. Take off preparation: head, shoulder, chest, hips, knees, feet must in the same direction. Body shape while airborne must be clearly recognizable. Body and legs must be tight and straight, with head in line with the spine. {\bf Landing Positions}: 1) Standing: Landing on one foot or two feet must be in a vertical position, with bend leg(s) before finishing in perfect alignment. 2) In push up: both hands and supporting feet must land at the same time in a controlled manner. 3) In split: must land from airborne phase to split form with both hands on each side of the body on the floor. 4) In frontal split: must land from airborne phase to frontal split form, both hands in front of the body.
\begin{itemize}
    \item {\bf Straight jump}: The body is in extended alignment, the pelvis is fixed – 2 different kinds of jumps and leaps: 1) Vertical: All air turns, Free fall. 2) Vertical to Horizontal: Gainer.
    \item {\bf Bent leg(s) jump}: 1) Tuck: Both legs are lifted close to the chest with knees bent. 2) Cossack: After takeoff, the body shows a pike shape, legs together parallel to the floor or higher, one leg straight, one leg bent. The angle between the trunk and legs: not be more than 60°. The angle at the knee joint may not be more than 60°.
    \item {\bf Pike jump}: After takeoff, the body shows a pike shape, legs together and straight, parallel to the floor or higher. The angle between the trunk and legs may not be more than 60°. 
    \item {\bf Straddle jump}: 1) Straddle: Legs are lifted in straddle position (minimum 90° angle), parallel to the floor or higher, arms and trunk extended over them.  The angle between the trunk and legs may not be more than 60°. 2) Frontal split:  Legs are fully abducted laterally (right and left) frontal (180°) with the upright upper body.
    \item {\bf Split jump}: 1) Split: Legs are fully stretched front and back in sagittal split (180°) with the upright upper body. 2) Switch: After takeoff, the leading leg must be parallel to the floor and switch with the rear leg to show a split (180°) in the air. 
    \item {\bf Scissors leap}: The leading leg must be parallel to the floor and switches forward with 1/2 turn (180°). 
    \item {\bf Kick jump}: The leading leg must be parallel to the floor and switches forward.
    \item {\bf Off axis jump}: A one-foot take off, kicking the free leg (bend or straight) upward diagonally. While airborne, the body inclines backward to be out of axis with a longitudinal rotation(s) in tuck or straight position, arms close to the chest. Landing in 1 foot/feet together or in split. 
    \item {\bf Butterfly jump}: A one-foot take off, kicking the free leg backward to lift the body upward. While airborne, legs fly open in straddle (or feet together) with the body in a horizontal position (with or without longitudinal rotation(s).  Landing on one leg. 
\end{itemize}

{\bf Group D: Balance \& Flexibility.}
\begin{itemize}
    \item {\bf Split}: Legs must be straight, in line, showing 180°. In Vertical Split: supporting leg must be in vertical position.
    \item {\bf Turn}: All exercises requiring turns must demonstrate complete rotations on the ball of the foot. Turns are completed when the heel of the turning foot touches the floor. 
    \item {\bf Balance turn}: A Balance turn where one leg is lifted to either in sagital or frontal balance and is supported by one hand.
    \item {\bf Illusion}: Starting position of illusion: head, shoulder, chest, hips, knees, toes must be in alignment. A full split (180°) must be shown during the movement. 
\end{itemize}

For the temporal definition, strictly follow the diagrams in Figure~\ref{fig:aerobic_demo} (quoted from \cite{de2017aerobic}) to determine the starting and ending of actions, except for the following situations: when an athlete’s action is not in place or is completely blocked by other athletes.

\subsection*{E.2 Volleyball}

\begin{itemize}
    \item {\bf Serve}: Send the ball over the net from behind the end line to start a new round. \textbf{Start}: The ball leaves the player's hand. \textbf {End}: If the player takes off, any foot touches the ground. Otherwise, the upper arm of the serving arm is below the horizontal plane.
    \item {\bf Block}: Deflect the ball coming from an attacker on the net. The one that doesn't take off is not considered a block. The one that takes off but doesn't touch the ball is considered a block. \textbf{Start}: Any foot leaves the ground. \textbf {End}: Any foot touches the ground.
    \item {\bf First Hit Pass}: Receive the serve. The player can receive the ball overhand, one-hand or underhand. \textbf{Start}: If overhand, the player raises any hand over the chest. If one-hand, the player's arm begins to stretch out. If underhand, the player begins to hold hands together. \textbf {End}: If overhand, the player puts any hand below the chest. If one-hand, the player's hitting-ball arm relaxes. If underhand, the player's hands loose.
    \item {\bf Defend}: Receive the ball from the opposite side except for the serve. The player can receive the ball overhand, one-hand or underhand. \textbf{Start}: If overhand, the player raises any hand over the chest. If one-hand, the player's arm begins to stretch out. If underhand, the player begins to hold hands together. \textbf {End}: If overhand, the player puts any hand below the chest. If one-hand, the player's hitting-ball arm relaxes. If underhand, the player's hands loose.
    \item {\bf Protect}: Receive the ball returned by the block. The player can receive the ball overhand, one-hand or underhand. \textbf{Start}: If overhand, the player raises any hand over the chest. If one-hand, the player's arm begins to stretch out. If underhand, the player begins to hold hands together. \textbf {End}: If overhand, the player puts any hand below the chest. If one-hand, the player's hitting-ball arm relaxes. If underhand, the player's hands loose.
    \item {\bf Second Hit Pass}: The second overhand pass to organize the offense. \textbf{Start}: The player raises any hand over the chest. \textbf {End}: The player puts any hand below the chest.
    \item {\bf Adjust}: For the second touch, due to the inadequacy of first hit pass or defending or protecting, the player has to adjust the ball underhand to organize offense. \textbf{Start}: The player begins to hold hands together. \textbf {End}: The player's hands loose.
    \item {\bf Save}: Due to the poor first hit pass or defending or protecting, the route of the ball changes dramatically. The actor can’t second hit pass overhand or adjust underhand to organize offense but uses one hand or both hands to reach the ball to prevent the ball from landing directly. \textbf{Start}: If one-hand, the player's arm begins to stretch out. Otherwise, the player begins to hold hands together. \textbf {End}: If one-hand, the player's hitting-ball arm relaxes. Otherwise, the player's hands loose.
    \item {\bf Second Attack}: For the second touch, a direct attack by the setter. \textbf{Start}: Any foot leaves the ground. \textbf {End}: Any foot touches the ground.
    \item {\bf Spike}: Slam the ball over the net into the opposing court. \textbf{Start}: Any foot leaves the ground. \textbf {End}: Any foot touches the ground.
    \item {\bf Dink}: Lightly tap the ball over the net to an area on the opponent’s side of the court that is not being guarded or occupied by a defensive player. \textbf{Start}: Any foot leaves the ground. \textbf {End}: Any foot touches the ground.
    \item {\bf No Offensive Attack}: For the second or third touch, the ball is passed over the net non-aggressively, because of the bad first hit pass or defending or protecting. The actor can push the ball overhand, pass the ball underhand or tap the ball from a position below the net with one hand, where the actor doesn't take off. \textbf{Start}: If overhand, the player raises any hand over the chest. If underhand, the player begins to hold hands together. Otherwise, the upper arm of hitting the ball arm is above the horizontal plane. \textbf {End}: If overhand, the player puts any hand below the chest. If underhand, the player's hands loose. Otherwise, the upper arm of hitting the ball arm is below the horizontal plane.
\end{itemize}
\subsection*{E.3 Football}

\begin{itemize}
    \item {\bf Shoot}: Hit the ball in an attempt to score a goal. Feet, torso and head are all allowed. \textbf{Start}: If using torso and head: if the player takes off, any part of the body leaves the ground (such as a foot); if the player does not take off, the player stands firmly and prepares to touch the ball. If using feet, the ball-controlling foot leaves the ground. \textbf {End}: If using torso and head: if the player takes off, any part of the body touches the ground (such as a foot); if the player does not take off, stand firmly after touching the ball. If using feet, the ball-controlling foot touches the ground.
    \item {\bf Long Ball}: Middle and long distance (over 30 meters) pass. Feet, torso and head are all allowed. \textbf{Start}: If using torso and head: if the player takes off, any part of the body leaves the ground (such as a foot); if the player does not take off, the player stands firmly and prepares to touch the ball. If using feet, the ball-controlling foot leaves the ground. \textbf {End}: If using torso and head: if the player takes off, any part of the body touches the ground (such as a foot); if the player does not take off, stand firmly after touching the ball. If using feet, the ball-controlling foot touches the ground.
    \item {\bf Pass}: Short distance (within 30 meters) pass. Feet, torso and head are all allowed. \textbf{Start}: If using torso and head: if the player takes off, any part of the body leaves the ground (such as a foot); if the player does not take off, the player stands firmly and prepares to touch the ball. If using feet, the ball-controlling foot leaves the ground. \textbf {End}: If using torso and head: if the player takes off, any part of the body touches the ground (such as a foot); if the player does not take off, stand firmly after touching the ball. If using feet, the ball-controlling foot touches the ground.
    \item {\bf Through Ball}: A pass that can clearly break through the opponent's line of defense and has a penetrating effect. At least one defensive player is passed. The ball is passed in front of the player's teammate. In other words, the player should pass the ball to where his running teammate is going to be. Feet, torso and head are all allowed. \textbf{Start}: If using torso and head: if the player takes off, any part of the body leaves the ground (such as a foot); if the player does not take off, the player stands firmly and prepares to touch the ball. If using feet, the ball-controlling foot leaves the ground. \textbf {End}: If using torso and head: if the player takes off, any part of the body touches the ground (such as a foot); if the player does not take off, stand firmly after touching the ball. If using feet, the ball-controlling foot touches the ground.
    \item {\bf Cross}: A medium-to-long-range pass from a wide area of the field towards the centre of the field near the opponent's goal. Provide direct or indirect shooting opportunities for offensive players. Feet, torso and head are all allowed. \textbf{Start}: If using torso and head: if the player takes off, any part of the body leaves the ground (such as a foot); if the player does not take off, the player stands firmly and prepares to touch the ball. If using feet, the ball-controlling foot leaves the ground. \textbf {End}: If using torso and head: if the player takes off, any part of the body touches the ground (such as a foot); if the player does not take off, stand firmly after touching the ball. If using feet, the ball-controlling foot touches the ground.
    \item {\bf Dribble}: Have control over the ball for a period of time and distance. \textbf{Start}: At the first touch with the ball, the ball-controlling foot leaves the ground. \textbf {End}: At the last touch with the ball, the ball-controlling foot touches the ground.
    \item {\bf Trap}: Use effective parts of the body to adjust the ball, including the speed and position of the ball. Feet, torso and head are all allowed. \textbf{Start}: If using torso and head: if the player takes off, any part of the body leaves the ground (such as a foot); if the player does not take off, the player stands firmly and prepares to touch the ball. If using feet, the ball-controlling foot leaves the ground. \textbf {End}: If using torso and head: if the player takes off, any part of the body touches the ground (such as a foot); if the player does not take off, stand firmly after touching the ball. If using feet, the ball-controlling foot touches the ground.
    \item {\bf Throw}: The player throws the ball from out of the field and the goalkeeper throws the ball. \textbf{Start}: Upper arms swing forward. \textbf {End}: Upper arms are below the horizontal plane.
    \item {\bf Save}: The goalkeeper uses his body parts (except his feet) to destroy the ball that is threatening to the goal. \textbf{Start}: After the ball is shot, the goalkeeper begins to move. \textbf {End}: Any part of the body touches the ground.
    \item {\bf Interception}: The defensive player consciously destroys the ball on the opponent's pass route. Feet, torso and head are all allowed. \textbf{Start}: If using torso and head: if the player takes off, any part of the body leaves the ground (such as a foot); if the player does not take off, the player stands firmly and prepares to touch the ball. If using feet, the interception foot leaves the ground. \textbf {End}: If using torso and head: if the player takes off, any part of the body touches the ground (such as a foot); if the player does not take off, stand firmly after touching the ball. If using feet, the interception foot touches the ground.
    \item {\bf Tackle}: The defensive player snatches the ball under the control of the offensive player. \textbf{Start}: the tackling foot leaves the ground. \textbf {End}: the tackling foot touches the ground.
    \item {\bf Clearance}: The defensive player destroys the ball in the backfield in order to gain the initiative in time and space. The main difference between clearance and long ball/ball is that long ball/ball aims at some player but clearance is aimless. Feet, torso and head are all allowed. \textbf{Start}: If using torso and head: if the player takes off, any part of the body leaves the ground (such as a foot); if the player does not take off, the player stands firmly and prepares to touch the ball. If using feet, the clearance foot leaves the ground. \textbf {End}: If using torso and head: if the player takes off, any part of the body touches the ground (such as a foot); if the player does not take off, stand firmly after touching the ball. If using feet, the clearance foot touches the ground.
    \item {\bf Block}: Intentionally destroy the opponent's threatening shot or block the opponent's shooting angle. The goalkeeper blocked the ball with his foot. Feet, torso and head are all allowed. \textbf{Start}: If using torso and head: if the player takes off, any part of the body leaves the ground (such as a foot); if the player does not take off, the player stands firmly and prepares to touch the ball. If using feet, the blocking foot leaves the ground. \textbf {End}: If using torso and head: if the player takes off, any part of the body touches the ground (such as a foot); if the player does not take off, stand firmly after touching the ball. If using feet, the blocking foot touches the ground.
    \item {\bf Defence}: The defensive player approaches the player, of whom the ball is under the control, to make restriction and interference. \textbf{Start}: The defender is shorter than 1.2 meters from the offensive player who is controlling the ball. \textbf {End}: 1) the offensive player passes the ball out or the ball is gained by other defensive players. 2) this defender begins to tackle. 3) this defender is longer than 1.2 meters from the offensive player who is controlling the ball.
    \item {\bf Aerial duels}: Two or more people compete for the high-altitude ball in order to obtain the ball, where all people are annotated. If the player does not take off, it is not considered aerial duels. Note that the player who has an obvious purpose, such as clearance and pass, is annotated that action. \textbf{Start}: Any part of the body leaves the ground (such as a foot). \textbf {End}: Any part of the body touches the ground (such as a foot).
\end{itemize}

\subsection*{E.4 Basketball}

\begin{itemize}
    \item {\bf Pass}: The player moves the ball to the teammate. \textbf{Start}: The player begins to push the ball outwards with his arms. \textbf{End}: The ball leaves both hands of the player.
    \item {\bf Drive}: The player, who controls the ball, gets rid of the defense by passing the defensive player or stopping suddenly during the movement. The aim is to get closer to the basket and create a space that is conducive to shooting. The next step is usually to shoot, layup, or pass the ball to teammates. \textbf{Start}: At the first touch with the ball, the hand presses the ball down. \textbf{End}: At the last touch with the ball, the ball is bounced into the hand.
    \item {\bf Dribble}: The player slaps the ball bounced from the ground continuously while on the spot or on the move. \textbf{Start}: At the first touch with the ball, the hand presses the ball down. \textbf{End}: At the last touch with the ball, the ball is bounced into the hand.
    \item {\bf 3-point Shot}: Shoot from beyond the three-point line. \textbf{Start}: The player raises the shooting hand over the chest. \textbf{End}: If the player takes off, any part of the body touches the ground (such as a foot). Otherwise, the player puts the shooting hand below the chest.
    \item {\bf 2-point Shot}: Shoot from within the three-point line. \textbf{Start}: The player raises the shooting hand over the chest. \textbf{End}: If the player takes off, any part of the body touches the ground (such as a foot). Otherwise, the player puts the shooting hand below the chest.
    \item {\bf Free Throw}: Unopposed attempts to score points by shooting from behind the free throw line. \textbf{Start}: The player raises the shooting hand over the chest. \textbf{End}: If the player takes off, any part of the body touches the ground (such as a foot). Otherwise, the player puts the shooting hand below the chest.
    \item {\bf Block}: When the offense shoots, the defender successfully knocks the ball out as the ball goes up. \textbf{Start}: The defender raises the blocking hand over the chest. \textbf{End}: If the defender takes off, any part of the body touches the ground (such as a foot). Otherwise, the defender puts the blocking hand below the chest.
    \item {\bf Offensive Rebound}: After a missed shot, the two sides compete for a rebound and the offensive player grabs it. \textbf{Start}: The player raises the grabbing-ball hand over the chest. \textbf{End}: If the player takes off, any part of the body touches the ground (such as a foot). Otherwise, the player catches the ball firmly.
    \item {\bf Defensive Rebound}: After a missed shot, the two sides compete for a rebound and the defensive player grabs it. \textbf{Start}: The player raises the grabbing-ball hand over the chest. \textbf{End}: If the player takes off, any part of the body touches the ground (such as a foot). Otherwise, the player catches the ball firmly.
    \item {\bf Pass Steal}: The defensive player intercept the ball in the process of passing, which is not under the control of offensive player. \textbf{Start}: The defender's stealing-ball hand begins to stretch out. \textbf{End}: 1) Route of the ball changes; 2) The defender catches the ball firmly.
    \item {\bf Dribble Steal}: The defensive player steals the ball under the control of offensive player. \textbf{Start}: The defender's stealing-ball hand begins to stretch out. \textbf{End}: The ball is out of the control of the offensive player who had the control.
    \item {\bf Interfere Shot}: The defender interferes with the shot but does not touch the ball. \textbf{Start}: The defender raises the interfering hand over the chest. \textbf{End}: If the defender takes off, any part of the body touches the ground (such as a foot). Otherwise, the defender puts the interfering hand below the chest.
    \item {\bf Pick-and-roll Defense}: In pick-and-roll, the defender of the offensive ball-controlling player is blocked by the teammate of this offensive player. \textbf{Start}: The defender has physical contact with the offensive screening player. \textbf{End}: The defender does not have physical contact with the offensive screening player.
    \item {\bf Sag}: The defender gives up the offensive player he is responsible for and turns to defend the offensive ball-controlling player. \textbf{Start}: The defender consciously approach the offensive ball-controlling player. \textbf{End}: 1) the offensive player passes or shoots the ball; 2) this defender is broken through; 3) this defender gives up.
    \item {\bf Screen}: In pick-and-roll, the offensive player uses his body to set a pick for his ball-controlling teammate. \textbf{Start}: Both feet of the offensive player touches the ground. \textbf{End}: Any foot of the offensive player is ready to leave the ground completely. Small range movement is not considered the end.
    \item {\bf Pass-inbound}: The player passes the ball from the boundary lines to restart the play. \textbf{Start}: The player begins to push the ball outwards with his arms. \textbf{End}: The ball leaves both hands of the player.
    \item {\bf Save}: The player gets back the ball that is about to go out of bounds. \textbf{Start}: The player begins to push the ball outwards with his arms. \textbf{End}: The ball leaves both hands of the player.
    \item {\bf Jump Ball}: A method used to begin or resume the play. Two opposing players attempt to gain control of the ball after an official tosses it into the air between them, where both players are annotated. \textbf{Start}: The player raises the grabbing-ball hand over the chest. \textbf{End}: Any part of the body touches the ground (such as a foot).
\end{itemize}

{\small
\bibliographystyle{ieee_fullname}
\bibliography{egbib}
}

\end{document}